\pdfoutput=1
\documentclass[11pt]{article}

\usepackage[final]{acl}

\usepackage{times}
\usepackage{latexsym}
\usepackage{amsmath}
\usepackage{amsfonts}
\usepackage{subcaption}
\usepackage{float}
\usepackage{cleveref}
\usepackage{graphicx}
\usepackage[T1]{fontenc}
\usepackage[utf8]{inputenc}

\usepackage{microtype}

\usepackage{inconsolata}

\usepackage{graphicx}
\usepackage{hyperref}

\title{Learning to vary: Teaching LMs to reproduce human linguistic variability in next-word prediction}
\author{
  Tobias Groot \\
  University of Amsterdam \\
  {\small \texttt{tobias.groot@student.uva.nl}} \\\And
  Salo Lacunes \\
  University of Amsterdam \\
  {\small\texttt{salo.lacunes@student.uva.nl}} \\\And
  Evgenia Ilia \\
  University of Amsterdam \\
  {\small \texttt{e.ilia@uva.nl}} \\
}

\begin{document}
\maketitle
\begin{abstract}
Natural language generation (NLG) tasks are often subject to inherent variability; \emph{e.g.} predicting the next word given a context has multiple valid responses, evident when asking multiple humans to complete the task. While having language models (LMs) that are aligned pluralistically, so that they are able to reproduce well the inherent diversity in perspectives of an entire population of interest is clearly beneficial, \citet{ilia2024predict} show that LMs do not reproduce this type of linguistic variability well. %. However, previous work shows that LMs do not perfectly reproduce the linguistic variability of a human population. Capturing this variability is crucial for generating more natural, human-like text and avoiding over-reliance on dominant linguistic patterns. 
They speculate this inability might stem from the lack of consistent training of LMs with data reflecting this type of inherent variability. As such, we investigate whether training LMs on multiple plausible word continuations per context  can improve their ability to reproduce human linguistic variability for next-word prediction. We employ fine-tuning techniques for pre-trained and instruction-tuned models; and demonstrate their potential when fine-tuning GPT-2 and Mistral-7B-IT, using Provo Corpus. %\citep{luke2018provo}. %, a next-word prediction dataset with multiple continuations per context. 
Our evaluation, which measures divergence among empirically estimated human and model next-word distributions across contexts before and after fine-tuning, %evaluate model outputs using Total Variation Distance (TVD) to quantify divergence from human response distributions. Our results demonstrate that 
shows that our multi-label fine-tuning improves the LMs' ability to reproduce linguistic variability; both for contexts that admit higher and lower variability. %as well as those with more restricted, high-consensus continuations.
\end{abstract}

\section{Introduction}

%An LM can be viewed as a representation of uncertainty about human linguistic production reflecting the production variability of the human population(s) that generated the training data.
%However, this production variability is not always perfectly with human response distributions \cite{ilia2024predict, pavlick2019inherent}. Previous work suspects that this misalignment stems from the fact that LMs are not consistently subjected to multiple responses per prompt during training. \citet{ma2025large} further analyze this issue, showing that LLMs generate multi-label predictions sequentially with overconfident, spiky distributions.

Inherent variability in natural language generation (NLG) tasks might arise from ambiguity or varying perspectives \citep{ plank2022problem, baan2023uncertainty}. For example, when predicting the next word given a context, multiple plausible and valid continuations exist; a task whose linguistic variability we can appreciate by asking a human population to complete it \citep{luke2018provo}.
%has seen rapid advancements in recent years, largely due to the success of
We can also appreciate this type of linguistic variability for autoregressive language models (LMs) that %are pre-trained on next-token prediction \citep{brown2020language, touvron2023llama, team2023gemini} and 
generate text by sampling from next-token (\emph{i.e.} subword unit) distributions %over a vocabulary 
conditioned on preceding tokens \citep{vaswani2017attention}. We achieve that by viewing such distributions as a representation of the model's uncertainty over continuations given a prefix \citep{ilia2024predict, guo2024benchmarking, tevet2020evaluating}. %, capturing the model's linguistic variability for the prefix .
% Whereas it is often valuable that models are able to reproduce such variability, as it makes for LMs that are more robust \citep{sheng2008get, peterson2019human, uma2021learning, kurniawan2025training} and representative of the linguistic diversity of human populations of interest \citep{sorensen2024roadmap, muscato2025perspectives}, 
% %However, this linguistic variability
% it has been shown that the variability LMs exhibit does not always align with the one humans exhibit \citep{ pavlick2019inherent, ma2025large, shaib2024detection}. For next word prediction, \citet{ilia2024predict} identify this misalignment and speculate it might stem from inconsistent exposure of LMs to training data reflecting such variability. 
It is often valuable for models to reproduce such variability, particularly in open-ended NLG tasks, where multiple responses can be plausible. %; reflecting the diversity in perspectives and beliefs of populations. 
Whereas this variability contributes to making LMs more robust \citep{sheng2008get, peterson2019human, uma2021learning, kurniawan2025training} and more representative of the linguistic diversity of human populations of interest \citep{sorensen2024roadmap, muscato2025perspectives}, it has been shown that the variability LMs exhibit does not always align with the one humans exhibit \citep{ pavlick2019inherent, ma2025large, shaib2024detection}. For next word prediction, \citet{ilia2024predict} identify this misalignment and speculate it might stem from inconsistent exposure of LMs to training data reflecting such variability. 

% discuss prior work that motivates improving human-like variability 
%Accurately capturing this variability is important because it enables language models to output more natural, human-like text that more accurately reflects the inherent uncertainty of language use, rather than exploiting the most likely continuation. 
%Moreover, capturing this variability has been shown to boost fairness in representing minority viewpoints \citep{muscato2025perspectives} and  through a technique known as multi-label fine-tuning, where models are fine-tuned using multiple plausible labels per data point. 

As such, we investigate whether training LMs with multiple  observations of the next word per context will improve their ability to reproduce human variability. 
While previous fine-tuning work utilising multiple references per instance focused on classification tasks \citep{peterson2019human, %he2018joint,
uma2021learning, rajeswar2022multi}, 
our work focuses on next-word prediction, a generative task. %\footnote{We view next-word prediction as a generative task, as words might be composed of several tokens.} 
Similar to \citet{eisape2020cloze}, who employ a form of multi-label distillation in next word prediction, we also employ a technique to fine-tune pre-trained LMs and extend  to instruction-tuned LMs. For the former, we alter the training signal, and for the latter we exploit a training data augmentation method to ensure that variability is observed.% during the fine-tuning stage. %implicitly uses this multi-label fine-tuning to improve human-like variability, but limits the fine-tuning to a relatively small LSTM model, which restricts the generalizability of their findings. Consequently, there remains an open question regarding the impact of multi-label fine-tuning on larger or instruction-tuned language models in generative tasks.
%This study aims to fill that gap by exploring whether fine-tuning of language models on multiple plausible continuations per context can enhance their ability to reproduce human-like linguistic variability. 
% Add section why it is even important to make models reproduce variability better. Then discuss about what you did in relation to other work (e.g. while other work did this and that, we look at the problem from this other lens, or improve upon the problem in this way). e.g. Mention here that you are tuning an instruction tuned model.

We employ these fine-tuning techniques for GPT-2 \citep{radford2019language}, a pre-trained model,  and Mistral-7B-IT \citep{jiang2023mistral7b}, an instruction-tuned model. When evaluating, by measuring divergence among empirically estimated human and
model next-word distributions across contexts, before and after fine-tuning, %whether the next-word distributions of the fine-tuned models align better with human next-word distributions of their non-fine-tuned counterparts.
% Findings
we find that fine-tuning with multiple labels per instance improves those LMs' ability to reproduce linguistic variability,  % for both PT and IT models. 
 across contexts of varying open-endedness. %, but also in more `restricted' contexts that admit less variability. %, demonstrating that the models learn to better capture the natural diversity of human language.
%Further analysis reveals that models fine-tuned with multiple labels predict a higher fraction of relevant unique words compared to their non fine-tuned counterparts.
Additional ablations %show that as few as 16 human labels are sufficient for substantial performance gains
measure performance when varying the number of training labels per instance; and with a preliminary analysis we measure the trade-off in performance in tasks that admit no plausible variability. For that, we handcraft a small evaluation dataset using a knowledge-based question answering dataset \citep{berant-etal-2013-semantic}.\footnote{Code available at: \href{https://github.com/tgroot56/Learning-to-vary-Teaching-LMs-to-reproduce-human-linguistic-variability-in-next-word-prediction}{GitHub repository}}

\section{Related Work}

% Human label variation in NLP tasks is often dismissed as annotator noise. However, \citet{plank2022problem} argue for a more nuanced interpretation, distinguishing between meaningless error and meaningful signal. They show that, particularly in tasks without a clear-cut correct answer, humans tend to produce distributions of plausible continuations rather than a single “gold” response. This kind of variability is not only expected, but observable in several datasets that include multiple annotations per item, revealing consistent and structured variability in human judgment \cite{webergenzel2024varierrnliseparatingannotation, nie2020can, luke2018provo}. These findings challenge the notion that disagreement is noise, and instead suggest that it encodes valuable linguistic insight.

Human label variation in natural language processing (NLP) tasks is often dismissed as noise \citep{paun2022statistical, ferracane2021did}. However, multiple responses can be plausible, %instead of a unique "gold" label response. This is 
especially relevant to ambiguous or open-ended tasks or prompts %, where genuine plausible disagreement in responses might exist 
\cite{plank2022problem, baan2023uncertainty, webergenzel2024varierrnliseparatingannotation, nie2020can, aroyo2015truth}. 
Embracing this plausible variation as part of NLP systems, which could make them more fair \citep{deng2023you, muscato2025perspectives} and robust \citep{peterson2019human, sheng2008get}, involves altering all stages of our systems' development pipelines: from dataset creation, collecting multiple labels per prompt \citep[][\emph{i.a.}]{luke2018provo, nie2020can}, to training, utilising these labels during the learning phase \citep[][\emph{i.a.}]{rodriguez2024federated, aroyo2012harnessing, padmakumar2024beyond}, and evaluation, comparing models' responses to multiple human references \citep[][\emph{i.a.}]{ baan2022stop, ilia2024predict}. 

Our approach aims to embrace plausible variability during training. Rather than collapsing annotations into a single ground truth \citep{paun2022statistical}, we incorporate multiple plausible references. % per input when training a model. 
The idea of multi-label fine-tuning has been adopted in image-classification \cite{peterson2019human, aurpa2024instructnet, rajeswar2022multi}, as well as in NLP, primarily for classification \citep{uma2021learning,jung2023cluster, he2018joint, bețianu2024dallmi, li2024culturellm, zhang2024regurgitative, li2025preservingdiversitysupervisedfinetuning, muscato2025embracing}. Additionally, recent efforts have applied instruction fine-tuning for multi-label text classification tasks \citep{siddiqui2024instruction, yin2024crisissense} and tasks with restricted outcome spaces, such as sampling from discrete distributions \citep{zhangforcing}.
Our work focuses on a generative task, (\emph{i.e.}, that of predicting complete wordforms by stringing together tokens), with a countably infinite outcome space (\emph{i.e.}, all possible wordforms from a finite set of tokens). \citet{eisape2020cloze} explores a form of multi-label distillation in next-word prediction for an LSTM model. We also explore a form of multi-label distillation for transformer-based models, extending our investigation to instruction tuned LMs. %, while retaining a probabilistic uncertainty lens. %this gap by fine-tuning two language models: GPT-2 \cite{radford2019language}, a base LM, and Mistral-7B \cite{jiang2023mistral7b}, an instruction-tuned model, on a multi-label next-word prediction dataset containing empirically obtained distributions of human responses. 
%By training these models explicitly to reflect the full distribution of human responses, we aim to better align their predictive variability with that of humans.

% introduce the gap here -> no multi-label training for generative tasks + not done for instruction tuned models 

%Our work simultaneously joins a stream of work that aims %for better alignment with diverse human preferences; for example, via reward regularization of reward models \citep{padmakumar2024beyond} or personalised alignment during inference \citep{chen2024pad}

\section{Methodology}
We exploit simple yet intuitive fine-tuning techniques, depending on the LMs' previous training. These require a set of contexts $C = \{c_1, ..., c_N\}$, where for each context $c_i$, we have a set of human next-word references $W_i = \{w_{i1}, ..., w_{iM}\}$\footnote{M might vary accross contexts.}:

\paragraph{Fine-tuning pre-trained LMs}\label{fine-tuning losses}
Autoregressive LMs are trained using cross-entropy between a target and the model’s next-token distribution given context $c$ (\( p(\cdot | c) \)  and \( q(\cdot | c) \) resp.). This corresponds to searching for the maximum likelihood estimate (MLE). 
%:
%\begin{equation}
%L_{\text{CE}} = -\sum_{w \in \mathcal{V}} p(w \mid c)\, \log q(w \mid c),
%\end{equation}
When training on a corpus with a single continuation (\emph{i.e.} the next corpus token $w^*$), \( p \) is a deterministic distribution centered on \( w^* \), leading to the following loss: %(\emph{i.e.} the token following the context in the corpus)
%, the target \( p \) is a deterministic distribution centered on the single next token \( w^* \) (the one found in the corpus); leading to the following loss:
\begin{equation}
L_{\text{Label}} = -\log q(w^* \mid c).
\end{equation}
%This setup encourages the model to concentrate probability mass on the observed next word, often resulting in overconfident and overly narrow predictions that ignore the natural variability present in human language.

When multiple word continuations are available, we replace this deterministic distribution with the empirically estimated distribution (using $W_i$), where the probability of a word given $c_i$,  \( p(w | c_i) \), equals its relative frequency in $W_i$. This results in the following loss, which comprises generalized cross entropy \citep{jm3}:
\begin{equation}
L_{\text{Var}} = -\sum_{w \in \mathcal{V}} p(w \mid c_i)\, \log q(w \mid c_i),
\end{equation}
where \( \mathcal{V} \) is the vocabulary.%, \( w \) is a word within the vocabulary, and \( c \) is a context.
\footnote{Words that actually contribute to the loss, i.e. non-zero terms, are words in the set of human samples, $W_i$ for $c_i$.}
%where \( \mathcal{V} \) is again the model vocabulary, however, now there are multiple words $w$ with non-zero terms contributing to the loss, namely the set of distinct next words observed for context $c$. Here, \( p(w \mid c) \) is computed as the relative frequency of \( w \) among all human continuations for \( c \). 
Since words may consist of multiple tokens, to obtain \( q(w \mid c_i) \) we re-express the model’s token-level probabilities over complete words.\footnote{Humans predicted \emph{word} continuations, not tokens; so the outcome space of \( p(w \mid c_i) \) is over complete words, and we must ensure that \( q(w \mid c_i) \) is expressed over the same space.} For a word $w$ with tokenization $\tau(w) = (t_1, \dots, t_n)$, we compute:
\begin{equation}
q(w \mid c_i) = \prod_{j=1}^n q(t_j \mid c_i, t_1, \dots, t_{j-1}),
\end{equation}
where \( q(t_j \mid \cdot) \) is the probability of token \( t_j \) under the model, given the context and preceding tokens.

% ; so we need to re-express the model's probabilities from the token to the complete word space)\footnote{Humans predicted \emph{word} continuations, not tokens; so the outcome space of \( p(w | c) \) is over complete words; we need to ensure \( q(w | c) \) is expressed over the same space.}, 
% % we decompose \( w \) into its tokenization and use the chain rule; details in Appendix \ref{app:model_probs}.
% \( \tau(w) = (t_1, \dots, t_n) \), and compute:
% \begin{equation}
% q(w \mid c) = \prod_{i=1}^n q(t_i \mid c, t_1, \dots, t_{i-1}),
% \end{equation}
% where \( q(t_i \mid \cdot) \) is the probability of token \( t_i \) under the model, given the context and preceding tokens.% This differs from the implementation of \citet{eisape2020cloze}, that considers probabilities in the space of tokens (and then uses it as a component of a final loss).

% while \( q(w \mid c) \) is approximated by sampling multiple continuations from the model.

% This objective encourages the model to align its output distribution with human variability by distributing probability mass across multiple plausible continuations, referred to as 'soft targets'.

\paragraph{Fine-tuning instruction-tuned LMs}\label{Fine-tuning instruct.}
Instruction-tuned models underwent additional training to cater a rather conversational format %between the user and the model 
and adhere to task instructions. %as expressed by users. 
We sample responses from the model's conditional predictive distribution (CPD) given a prompt, \emph{i.e.} an instruction and an example. For our task, we sample response $r$ containing a predicted word from $q(r|(I,c_i))$, where the prompt includes instruction $I$ requesting a word continuation given a prefix, and the example context $c_i$. 
% \begin{equation}
% q(r|p=(I,c))
% \end{equation}
%where $p$ is the prompt and , containing the predicted word, $w$. 
So as to utilise multiple labels, we employ the following training data augmentation technique: for each context $c_i$ in $C$, we construct the prompt $(I,c_i)$ and for each word $w_j$ in $W_i$, we create a training datapoint where $w_j$ is a response to $(I,c_i)$. This entails that $c_i$ will appear multiple times with different continuations as per their frequency in $W_i$. We train using $L_{Label}$.\footnote{Constructing the dataset in this way (one prompt-response pair for each word annotation for every context) using $L_{Label}$ is similar to learning $q(r|(I,c))$  with $L_{Var}$. } See Appendx \ref{App:one-shot} for prompts.

\section{Experiments}
\paragraph{Models \& Datasets.}
We fine-tune pre-trained GPT-2 (124M; \citet{radford2019language}) and instruction-tuned Mistral-7B-Instruct-v0.3 (7.25B; \citet{jiang2023mistral7b})%\footnote{\url{https://huggingface.co/mistralai/Mistral-7B-Instruct-v0.3}}
, which we refer to as Mistral-7B-IT. 
Both models are fine-tuned using Provo Corpus \cite{luke2018provo}, which contains 55 text passages (2687 total contexts). Each prefix is annotated with an average of 40 human annotations predicting the word following it.
We split the dataset randomly at the paragraph level (to avoid partial passage leaks between train and test sets). 80\% is  for training, of which 10\% is reserved for validation; and the remaining for testing. %Splitting by paragraph avoids data leakage, as contexts from the same passage do not appear across splits.

\paragraph{Training Configuration.}
Both models were fine-tuned using the Adam optimizer \cite{kingma2014adam}. For GPT-2: we train for 3 epochs, using  a learning rate of $1e^{-5}$ and a batch size of 16. For Mistral-7b-IT: we train for 4 epochs using Low-Rank Adaptation \cite[][LoRA]{dettmers2023qloraefficientfinetuningquantized} with a learning rate of $1e^{-4}$ with a batch size of 32. %Quantization approximates a pre-trained model’s weights by representing them with a lower number of bits \cite{dettmers2023qloraefficientfinetuningquantized}. LoRA allows for efficient fine-tuning of LLMs because it leaves the model's original parameters frozen while adding a separate smaller set of new weights between attention layers, called LoRA adapaters, which are the only weights updated during fine-tuning \cite{hu2021loralowrankadaptationlarge}. 
We train on 3 random seeds; %(42, 123, 456). 
training details in Appendix \ref{App:qlora}.

% For exceptionally large language models, we made use of quantization in conjunction with Low-Rank Adaptation (LoRA). Quantization approximates a pre-trained model’s weights by representing them with a lower number of bits \cite{dettmers2023qloraefficientfinetuningquantized}. Ideally, quantization enables the model to operate at a lower precision level while preserving performance and significantly improving memory requirements.

% LoRA allows for efficient fine-tuning of LLMs because it leaves the model's original parameters frozen while adding a separate smaller set of new weights between attention layers, called LoRA adapaters, which are the only weights updated during fine-tuning \cite{hu2021loralowrankadaptationlarge}. These adapters can be seen as weights updates that the main model's frozen weights would require in order to succeed at the fine-tuned task--they are merged during inference. By only optimizing these low rank weights over the large full rank weights of the original model, we gain significant computational efficiency during fine-tuning.

\paragraph{Metrics.}\label{Metrics}
Following \citet{ilia2024predict}: for each context, we measure the divergence between the human and model CPDs given a context using total variation distance (TVD) \cite{rudin1987real}.\footnote{$\mathrm{TVD}(p, q) = \frac{1}{2}\sum_{w} |p(w|c_i)-q(w|c_i)|$}  TVD quantifies the difference between two probability distributions by summing the absolute differences in the probabilities they assign to the same event. A higher TVD indicates greater disagreement between human and model CPDs (\emph{i.e.}, poorer alignment with human linguistic variability), whereas a lower TVD indicates less disagreement (\emph{i.e.} better alignment with human linguistic variability).
In order to compute TVD, we need estimates of the human and model CPDs ($p(w|c)$ and $q(w|c)$ respectively). As done in \citet{ilia2024predict}: (1) for $p(w|c)$, we estimate it via Monte Carlo, with $p(w|c)$ equaling the relative frequency of $w$ in all human samples, and (2) for $q(w|c)$, we estimate it via Monte Carlo, by sampling 40 sequences from the model long enough to contain a full word, slice it, and compute $q(w|c)$ (or $q(w|(I,c))$) as the relative frequency of $w$ in all sampled words. 

\paragraph{Baselines \& Upper Bounds.} We compare the distribution of TVD values across contexts before and after fine-tuning, where improved performance would mean a shift towards lower TVD values (\emph{i.e.} less disagreement with human CPDs). For the instruction-tuned model, we add a 1-shot baseline, where the prompt includes an example of a context and word references (details in Appendix \ref{App:one-shot}). As another baseline, we fine tune models with Provo's original corpus passages (\emph{i.e.} one continuation per prefix), imitating models' usual training. %We also assess whether results we observe stem from an out-of-distribution effect of Provo Corpus compared with the models' training data.
Lastly, to estimate the best performance we can expect from our models, which essentially is to mimic human divergence, we establish a baseline for the expected level of disagreement from humans for a context. We split  human responses in two disjoint groups and measure their CPDs' TVD (`Oracle' baseline). %treating one as the target distribution and the other as a proxy for model output. 
%The resulting  `Oracle' TVD values provide a baseline for the amount of variation that naturally occurs among humans.

 \begin{table}
  \centering
  \scalebox{0.9}{
  \begin{tabular}{l cc}
    \hline
    \multicolumn{3}{c}{\textbf{Mean TVD $\pm$ SD ($\downarrow$)}} \\
    \textbf{Model} & \textbf{GPT-2} & \textbf{Mistral-7b-IT} \\
    \hline
    Base   & 0.607$\pm$0.001 & 0.812$\pm$0.002 \\
    1-Shot    & N/A & 0.784$\pm$0.002 \\
    FT (Orig. corpus) & 0.612$\pm$0.002 & 0.805$\pm$0.001 \\
    FT (Maj. label)   & 0.556$\pm$0.005 & 0.563$\pm$0.002 \\
    FT (Mul. labels)      & \textbf{0.550}$\pm$0.003 & \textbf{0.499}$\pm$0.006 \\
    \hline
    Oracle           & 0.443$\pm$0.002 & 0.443$\pm$0.002 \\
    \hline
  \end{tabular}}
  \caption{Mean and standard deviation of TVD averages across test contexts for three seeds.}
  \label{tab:tvd_results}
\end{table}

% \begin{figure*}[!h]
%   \centering
%   \begin{subfigure}[t]{0.48\textwidth}
%     \includegraphics[width=\linewidth]{imgs/checkthisout-2.pdf}
%     \caption{GPT-2}
%     \label{fig:tvd_main_gpt2}
%   \end{subfigure}
%   \hfill
%   \begin{subfigure}[t]{0.48\textwidth}
%     \includegraphics[width=\linewidth]{imgs/checkthisout-4.pdf}
%     \caption{mistral-7b-IT}
%     \label{fig:tvd_main_mistral}
%   \end{subfigure}
%   \caption{Distribution of TVD scores across contexts. For both GPT-2 and Mistral-7B-IT; fine-tuning shifts the TVD distribution toward the Oracle baseline, suggesting improved alignment with human linguistic variability.}
%   \label{fig:tvd_main_combined}
% \end{figure*}

% \begin{figure}[!h]
%   \centering
%   \begin{subfigure}[t]{0.48\textwidth}
%     \includegraphics[width=\linewidth]{imgs/checkthisout-2.pdf}
%     % \caption{GPT-2}
%     \label{fig:tvd_main_gpt2}
%   \end{subfigure}
%   \hfill
%   \begin{subfigure}[t]{0.48\textwidth}
%     \includegraphics[width=\linewidth]{imgs/checkthisout-4.pdf}
%     % \caption{mistral-7b-IT}
%     \label{fig:tvd_main_mistral}
%   \end{subfigure}
%   \caption{Distribution of TVD scores across contexts. For both GPT-2 and Mistral-7B-IT; fine-tuning shifts the TVD distribution toward the Oracle baseline, suggesting improved alignment with human linguistic variability.}
%   \label{fig:tvd_main_combined}
% \end{figure}

\begin{figure}[!ht]
  \centering
  \includegraphics[width=\linewidth]{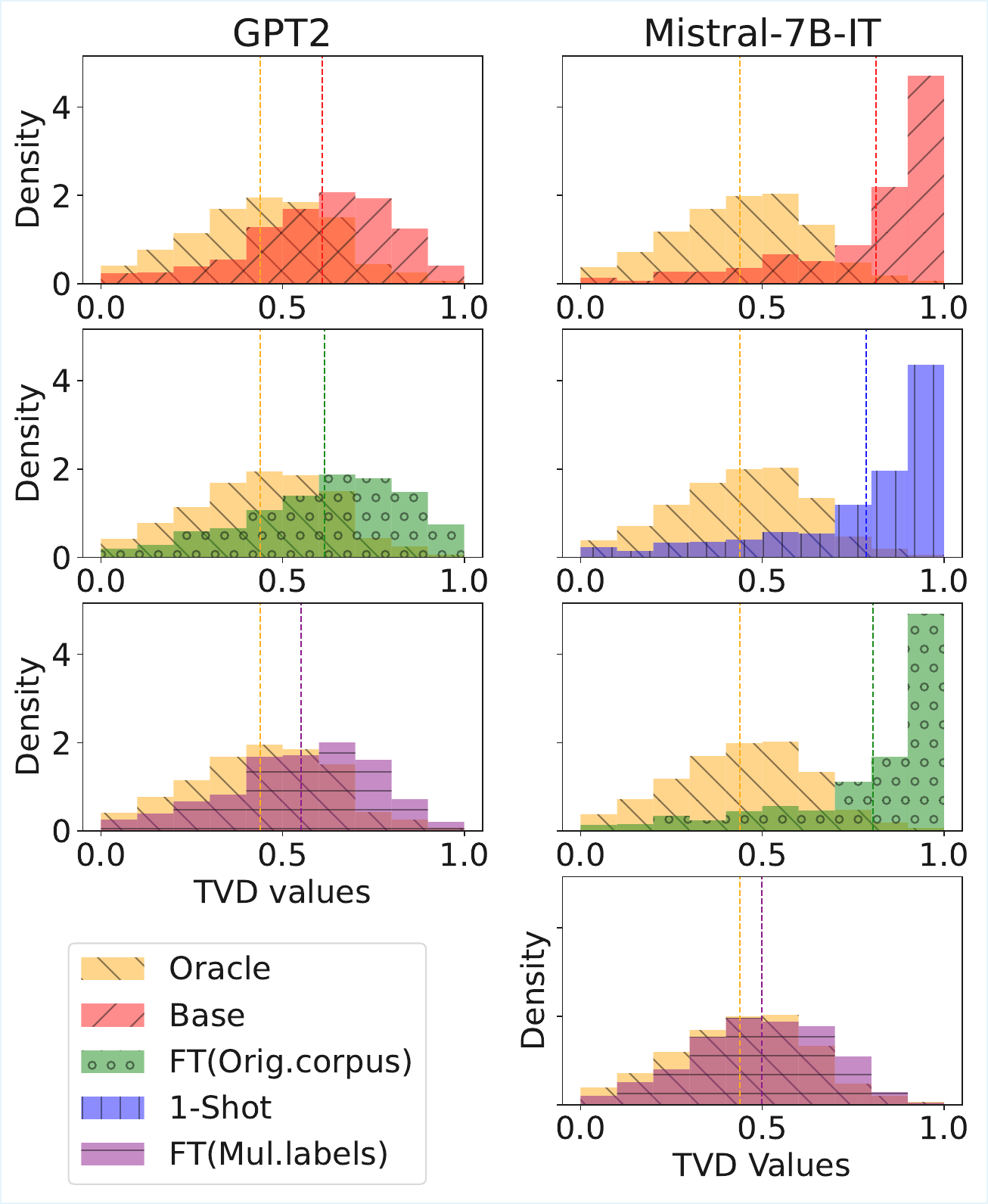}
    % \caption{GPT-2}
  \label{fig:tvd_main_gpt2}
  \hfill
  \caption{Distribution of TVD scores (for 1 seed) across contexts. For both GPT-2 and Mistral-7B-IT; fine-tuning shifts the TVD distribution towards the Oracle baseline, suggesting better linguistic alignment with humans.}
  \label{fig:tvd_main_combined}
\end{figure}

\section{Results}
\paragraph{Main results}
As shown in Table \ref{tab:tvd_results}, both models fine-tuned with multiple labels (FT (Mul. labels)) achieve a notably lower mean TVD compared to other baselines (Base, 1-Shot and FT (Orig. Corpus)). We also observe how FT (Orig. corpus)'s performance is very similar to the Base model. This simultaneously indicates that our improved performance does not stem from an out-of-distribution effect between Provo Corpus and the models' training data. %Fine-tuning the GPT-2 model reduces the TVD from 0.61 to 0.55, whereas Mistral-7B drops from 0.80 to 0.52. 
Figure \ref{fig:tvd_main_combined}, which shows the histogram of TVD values for all models and baselines (for 1 seed), confirms that; FT (Mul. labels) models' TVD distributions shift towards the Oracle distributions, indicating that models improve at reproducing human linguistic variability. For other seeds, we see similar patterns; see Appendix \ref{app:main_other_seeds}.

\paragraph{When and how do models improve?}To understand the effects of our fine-tuning, %with multiple human labels per instance
we analyze changes in TVD. We visualise the models' changes in performance against context open-endedness 
(as measured by the TVD between human oracles; lower TVD indicating more `restrictive' contexts), allowing us to grasp if performance gains arise in contexts that admit higher or lower variability. In Figure \ref{fig:differences_tvd_vs_tvdoracles} (Appendix \ref{app:anal_perform_changes}), negative TVD differences between fine-tuned and base models (indicating gains) occur at all levels of contexts' open-endedness.
%Figure \ref{fig:dominant_word_acc_combined} shows that TVD improvements for both fine-tuned models (indicated by blue dots and regression lines) are relatively consistent across different levels of human agreement on the dominant next word. Although both regression lines show a slight upward trend, indicating marginally greater improvements in high-consensus contexts, TVD gains are observed across both low- and high-consensus contexts. This suggests that the fine-tuned models' improved performance is not solely driven by more accurate prediction of the dominant word, but also reflects gains in contexts where human responses are more diverse. 
We also assess whether models improve at predicting words that humans predicted (regardless of frequency). We plot the fraction of unique human predictions that were also predicted by the models before and after fine-tuning (Figure~\ref{fig:unique_word_hist_combined}; Appendix \ref{app:anal_perform_changes}).
%we evaluate the coverage of unique relevant next word predictions, that is, the fraction of unique words predicted that match the unique words of human predictions. 
Fine-tuned Mistral-7B-IT's ability to predict unique human words (along with its CPDs' `diversity'; \Cref{fig:model_entropy_hist}, Appendix \ref{app:anal_perform_changes}) improves substantially %their non-fine tuned counterparts 
%; as we see a shift towards higher coverage levels post fine-tuning 
(details and analyses in Appendix \ref{app:anal_perform_changes}). %showing a clear rightward shift in the distribution of unique word coverage for the fine-tuned models for both GPT-2 and Mistral-7B. This indicates that the fine-tuned model predicts a greater number of relevant unique words per context compared to the non-fine-tuned baseline. 
%Together with the results on dominant word prediction, this suggests that the observed TVD improvements stem not only from better prediction of the most common human response, but also from an increased ability to capture the diversity of human-like next-word options.

\paragraph{Is the entire response distribution useful?}
When gathering datasets with multiple labels, disagreement can be discarded as noise and the most common response is used as ground truth. Aiming to assess whether retaining the entire response distribution %as opposed to retaining only the most popular response 
is useful, we fine-tune a model on Provo Corpus using only the majority response (FT (Maj. Label) in \Cref{tab:tvd_results}). %, and compare to FT (Mul. Labels). %our fine-tuned models that utilise the entire response distribution. 
We find that, FT (Maj. Label) surpasses the performance of FT (Orig. Corpus), which is not entirely surprising: the corpus word is a single observation, while the majority vote exploits in a sense multiple labels. %, optimising for a `pruned', deterministic distribution 
%(Figure~\ref{fig:model_entropy_hist}, Appendix \ref{app:anal_perform_changes}). 
%To assess whether the improvement in TVD is specifically attributable to the use of soft-target supervision rather than to fine-tuning in general, seperate models were fine-tuned using only one label as a hard target. Both the majority vote label and the original word present in the corpus are used as hard targets. This ablation isolates the effect of replacing the full human response distribution with a one-hot label representing the most commonly selected next word.
%In Table \ref{tab:tvd_results}, the mean TVD scores of the two separate hard target (HT) models are shown. Fine-tuning the models with the original corpus word does not improve the mean TVD scores compared to the non-fine-tuned models.
%Fine-tuning using the majority vote label as the hard target does, however, improve performance and achieves a similar overall improvement in TVD compared to the model fine-tuned with soft targets (ST), while the Mistral-7B ST model slightly outperforms its HT counterpart. 
This is intuitively in line with analysis revealing that performance gains seem to relate with less open-ended contexts (Figure~\ref{fig:differences_tvd_vs_tvdoracles_majlabel}; Appendix \ref{App:majlabelFT}).  %, as opposed to fine-tuning that exploited all available labels and improved across contexts of varying admitted variability.
%However, Figure~\ref{fig:dominant_word_acc_combined} reveals that this improvement is primarily concentrated in contexts with high human consensus, while gains are substantially smaller for low-consensus contexts, indicated by the steeper regression slope for the HT models (red crosses and lines) using majority vote labels compared to the ST models. 
%Moreover, Figure~\ref{fig:unique_word_hist_combined} shows that the distribution of relevant unique word predictions for the hard-target models is considerably worse compared to the soft-target models. For GPT-2, the model trained on hard targets even shifts to the left relative to the non-fine-tuned baseline. This indicates a reduction in lexical diversity, meaning that the model predicts fewer relevant unique words compared to the human distribution.
Nonetheless, FT (Mul. labels) outperforms FT (Maj. label), with  moderate gains for GPT2 and more notable gains for Mistral-7B-IT; indicating the utility of retaining all labels. %information relevant to label variation. %genuine and inherent disagreement when fine-tuning models for better reproducing human variability.

\paragraph{Number of labels ablation.}
We analyse how the number of labels used to fine-tune the model affect the model's performance. We fine-tune GPT2 using a varying number of labels each time (1,2,4,16 and 32; randomly sampled from available annotations). Figure~\ref{fig:ablation_samples} of Appendix \ref{App:ablation} shows that 16 samples are sufficient for substantial performance improvements; for more details, see Appendix \ref{App:ablation}.%; scores for 16 and 32 samples are nearly identical. Overall, gains are modest, ranging from 0.566 for two samples to 0.551 for 16 or more. These results suggest that while more human responses improve alignment with the empirical distribution, the marginal benefit diminishes, indicating that a limited number of high-quality annotations is sufficient to capture most human-like variability.

%\paragraph{Ablations.}
% For evaluation on tasks with a single correct answer, we manually adapted a subset of the WebQuestions dataset \cite{berant-etal-2013-semantic} into a next-word prediction format (examples are shown in Appendix \ref{App:dataset}). For evaluating the performance of the models on the task with one ground-truth label, the 'mean hit rate' will be used, which corresponds to the fraction of model samples where the gold target appears.

\paragraph{Impact on tasks without data uncertainty.}
Whereas optimising for a task that admits inherent variability (\emph{i.e.} next-word prediction) might improve the model's ability to reproduce such variability; the effect of this on tasks that admit no variability is unclear. To assess that, we test the models' performance before and after fine-tuning on knowledge-based question answering (a task admitting no plausible variability), adapted for next-word prediction. For that, we handcraft examples from a subset of WebQuestions \citep{berant-etal-2013-semantic}; details and examples in Appendix \ref{App:dataset}. For each context, we sample 40 responses and measure how often responses exactly match the reference. 
As shown in Table~\ref{tab:hit_rate_results} of Appendix \ref{App:dataset}, fine-tuning on multi-label data moderately improves the low performance of GPT2, but worsens the performance of Mistral-7B-IT; highlighting a potential trade-off in performance between tasks that do and do not admit variability, when optimising for the latter.%, the proportion of outputs matching the canonical label drops from 0.59 to 0.12. Interestingly, GPT-2 shows a slight increase from 0.03 to 0.04, although its overall performance remains poor. This may be attributed to GPT-2's limited model capacity and comparatively smaller knowledge base, which hinders its ability to retain factual information. The performance drop observed for Mistral-7B aligns with the increased output variability illustrated in Figure~\ref{fig:entropy}, highlighting the trade-off between diversity and precision in tasks that rely on a single correct answer. For a more detailed analysis, see Appendix~\ref{App:RQ3}.

\section{Conclusion}
This study examines whether fine-tuning with multiple labels per instance has the potential to enhance models' ability to reproduce linguistic variability in next word prediction. We show improved performance for a smaller pre-trained language model (GPT-2) and a larger instruction-tuned model (Mistral-7b-IT) across contexts that admit varying levels of plausible variability. Our findings highlight both the potential and possible limitations of such fine-tuning, paving the way for further advancements in modeling linguistic variation.
%Further analysis reveals that these improvements are not limited to high-consensus contexts but also extend to cases where human responses are more diverse, suggesting that soft-target supervision supports a more generalizable form of variability modeling. Moreover, we find that this benefit comes with a trade-off: a measurable drop in performance on tasks that require prediction of a single correct label.

\section{Limitations}
We hereby discuss various limitations of our study: we fine-tune using Provo Corpus, which is a relatively small dataset with a limited number of human annotations per prefix. The high cost of obtaining data with multiple references means that such data is scarce and not available at large scale. However, we show that even with a limited amount of contexts and a limited amount of annotations per context that are well-curated and of high-quality it is possible to observe performance improvements. Simultaneously, as the field of synthetic data generations is becoming increasingly popular; we can entertain the idea that future work exploits such synthetic labels, and a model that has been fine-tuned to embrace variability, such as the ones we present in this study, could comprise generators for such synthetic annotations.
Additionally, for our training and evaluation, we assumed all human annotations to be draws from the same underlying distribution; which is not an assumption that is easy to verify. 
We also observed a trade-off between capturing variability well and performance on tasks with a single correct answer; with future work potentially focusing on methods that could balance-off better such trade-offs. % While multi-label fine-tuning improved alignment with human-like variability, it led to a measurable drop in accuracy on deterministic next-word prediction tasks. This highlights a limitation of the approach in settings where precision or factual correctness is essential. This inspires future work exploring methods that could potentially balance better such trade-offs.
Additionally, due to resource constraints, we were only able to include in our study only two (relatively small) models that were trained for English. Despite focusing on a generative task, we only focused on next word prediction. Transferring this to the sequence level might be non-trivial and come with its own challenges. However, we hope that our study inspires future work in this research direction, aiming to embrace inherent variability as part of the training of LMs, and tackle challenges related to this field. 
%Furthermore, while the results on GPT-2 and Mistral-7B are promising, current state-of-the-art language models are significantly larger and more complex than those used in this study. Generalization of our findings to these larger models is not ensured, as differences in scale, and architecture may lead to different responses to multi-label fine-tuning.

\section*{Acknowledgements}
Evgenia Ilia is supported by the EU’s Horizon Europe research and innovation programme (grant agreement No. 101070631, UTTER). The experiments and findings presented in this paper were conducted as part of a research project fostered within the NLP 2 course of the MSc AI programme of the University of Amsterdam (2024-2025 edition), coordinated by Ana Lucic.

% \bibliography{custom}

\appendix

% \section{Model probabilities under tokenisation}
% \label{app:model_probs}
% To obtain \( q(w \mid c) \) (where words might be composed by several tokens; so we need to re-express the model's probabilities from the token to the complete word space)\footnote{Humans predicted \emph{word} continuations, not tokens; so the outcome space of \( p(w | c) \) is over complete words; we need to ensure \( q(w | c) \) is expressed over the same space.}, we decompose \( w \) into its tokenization \( \tau(w) = (t_1, \dots, t_n) \), and compute:
% \begin{equation}
% q(w \mid c) = \prod_{i=1}^n q(t_i \mid c, t_1, \dots, t_{i-1}),
% \end{equation}
% where \( q(t_i \mid \cdot) \) is the probability of token \( t_i \) under the model, given the context and preceding tokens.

\section{Prompts for Baselines}
\label{App:one-shot}
When constructing the training set and evaluating our models, we present the relevant prompts:

\paragraph{Base prompt.}
To assess the performance of the non fine-tuned models, we prompt them repeatedly for the next word prediction task. The prompt includes an instruction to predict a next-word continuation and the given context at a time. 

\textbf{Prompt:}

\begin{verbatim}
    Instruction: Return one plausible next 
    word for the following context. 
    Context: <CONTEXT>
    Continuation: 
\end{verbatim}

When creating training prompt-response pairs, the prompt is identical to before, and the responses are words from the set of human references.

\textbf{Response:}
\begin{verbatim}
     <HUMAN_REFERENCE>
\end{verbatim}

\paragraph{1-Shot prompt.}
As a performance baseline we  have one-shot prompting, which includes the instruction, an example from the training set, and the given context at a time:
\begin{verbatim}
    Instruction: This is an example of a 
    context and some plausible next word 
    continuations. given by a group of 39
    people: Context: There are now 
    rumblings that, Continuations:
    [are, are, are, are, are, are, are,
    can, can, can, can, can, sound,
    sound, sound, sound, shake, shake,
    shake, the, the, the, have, have,
    our, our, someone, someone, 
    appear, ca, cause, come, make,
    occur, people, say, suggest,
    tumble, we]. Following this
    example, return only one plausible
     next word for the following context.
    Context: <context>
    Continuation:
\end{verbatim}

\section{QLoRa}\label{App:qlora}
Table~\ref{tab:lora_bnb_config} shows the configuration used for finetuning the Mistral-7B-IT model. 
\begin{table}[!ht]
\footnotesize
\centering
\begin{tabular}{ll}
\hline
\textbf{Parameter} & \textbf{Value} \\
\hline
\multicolumn{2}{l}{\textbf{QLoRA}} \\
\hline
$r$ & 8 \\
LoRA $\alpha$ & 16 \\
LoRA dropout & 0.05 \\
Task type & Causal Language Modeling \\
Target modules & \texttt{q\_proj}, \texttt{k\_proj}, \texttt{v\_proj}, \texttt{o\_proj}, \\& \texttt{gate\_proj}, \texttt{up\_proj}, \texttt{down\_proj} \\
\hline
\multicolumn{2}{l}{\textbf{Quantization}} \\
\hline
Load in 4-bit & True \\
4-bit quantization type & nf4 \\
Double quantization & True \\
Compute data type & bfloat16 \\
\hline
\end{tabular}
\caption{LoRA and 4-bit quantization configuration parameters.}
\label{tab:lora_bnb_config}
\end{table}

\section{Main results}
\label{app:main_other_seeds}
\begin{figure*}[!ht]
  \centering
  \begin{subfigure}[t]{0.48\textwidth}
    \includegraphics[width=\linewidth]{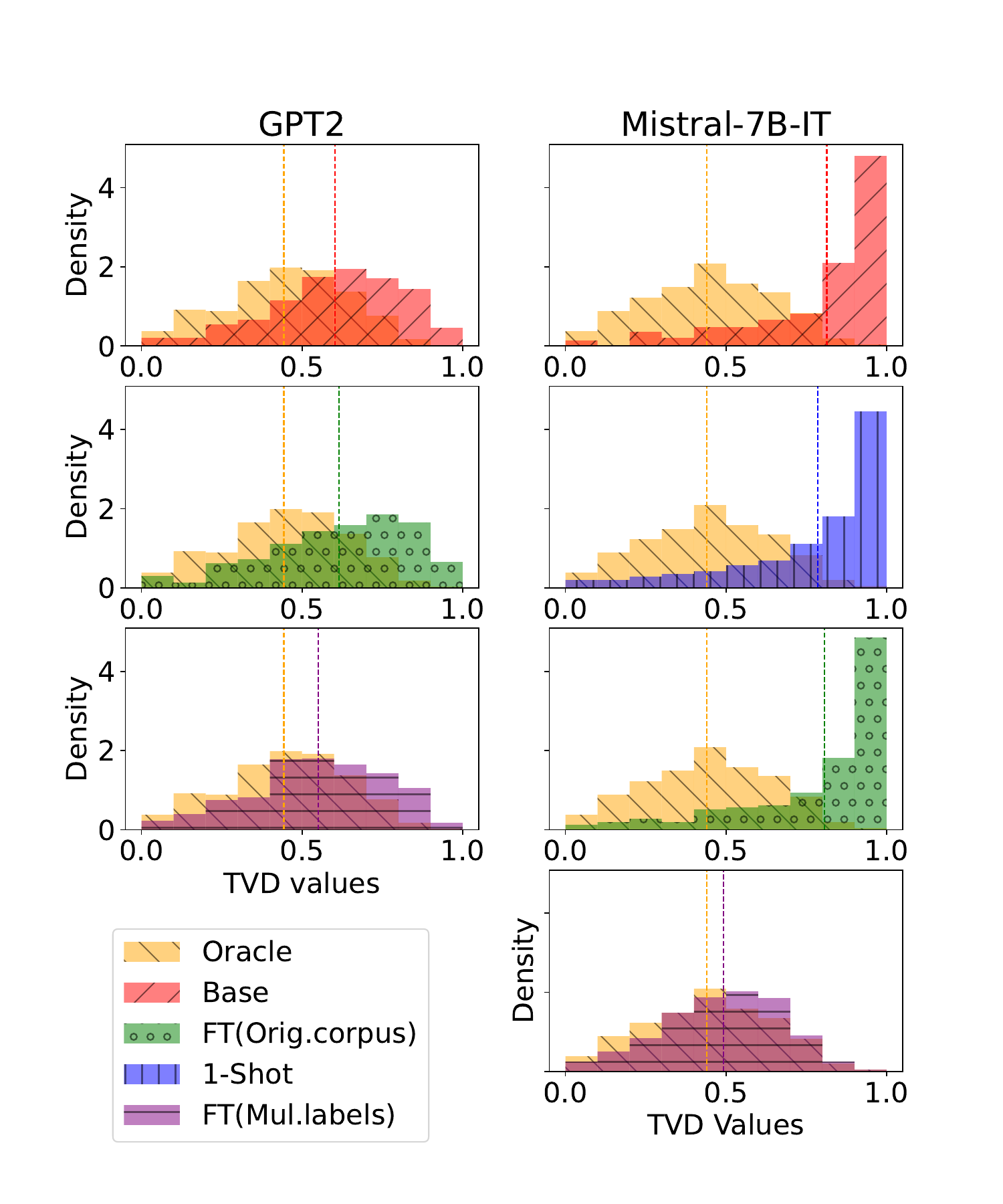}
    % \caption{GPT-2}
    \label{fig:tvd_main_seed1}
  \end{subfigure}
  \hfill
  \begin{subfigure}[t]{0.48\textwidth}
    \includegraphics[width=\linewidth]{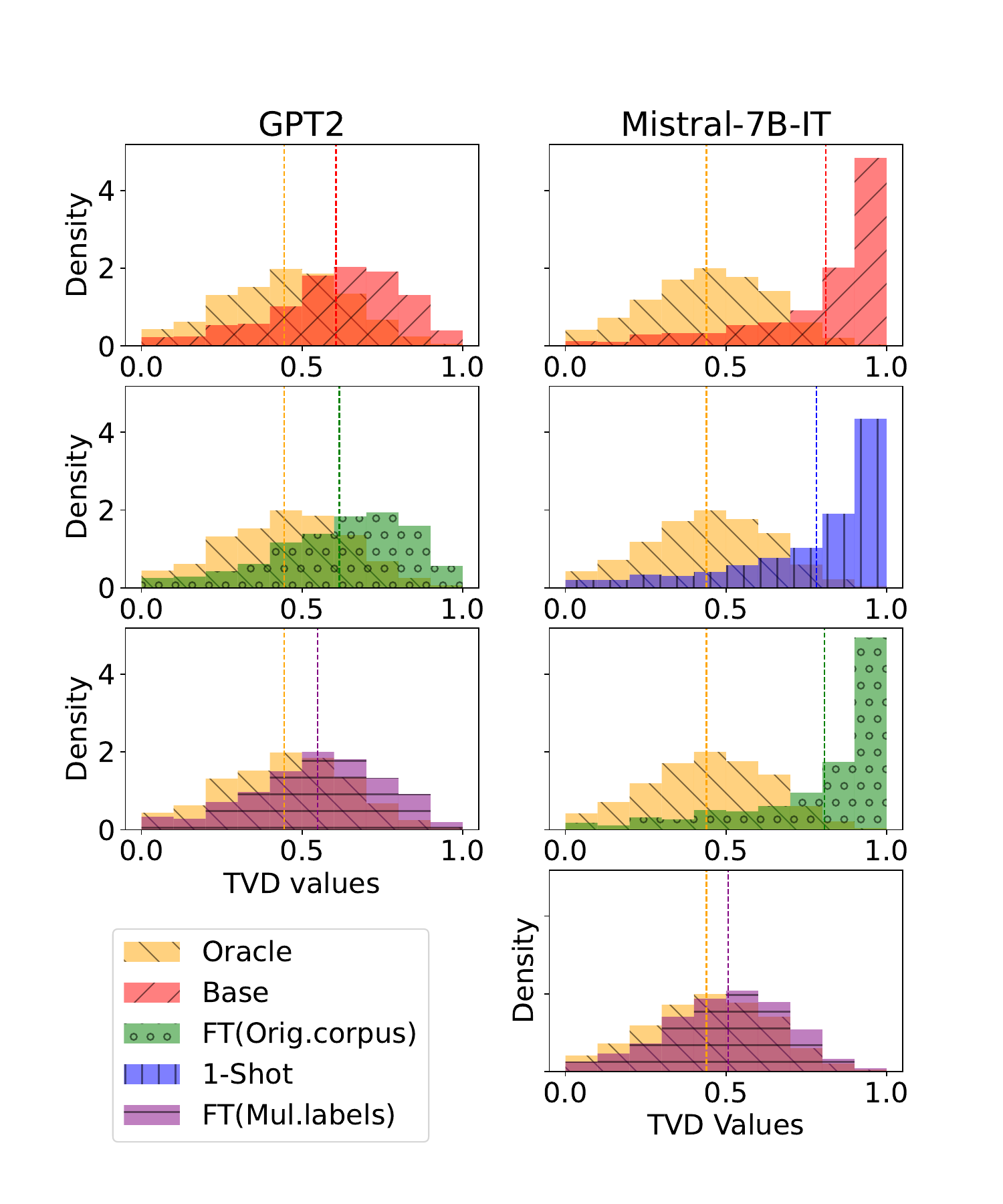}
    % \caption{mistral-7b-IT}
    \label{fig:tvd_main_mistral}
  \end{subfigure}
  \caption{Distribution of TVD scores across contexts, for the two remaining seeds not presented in the main paper. For both GPT-2 and Mistral-7B-IT; fine-tuning shifts the TVD distribution toward the Oracle baseline, suggesting improved alignment with human linguistic variability.}
  \label{fig:tvd_main_seed2}
\end{figure*}

\begin{figure*}[!ht]
  \centering
  \begin{subfigure}[t]{0.3\textwidth}
    \includegraphics[width=\linewidth]{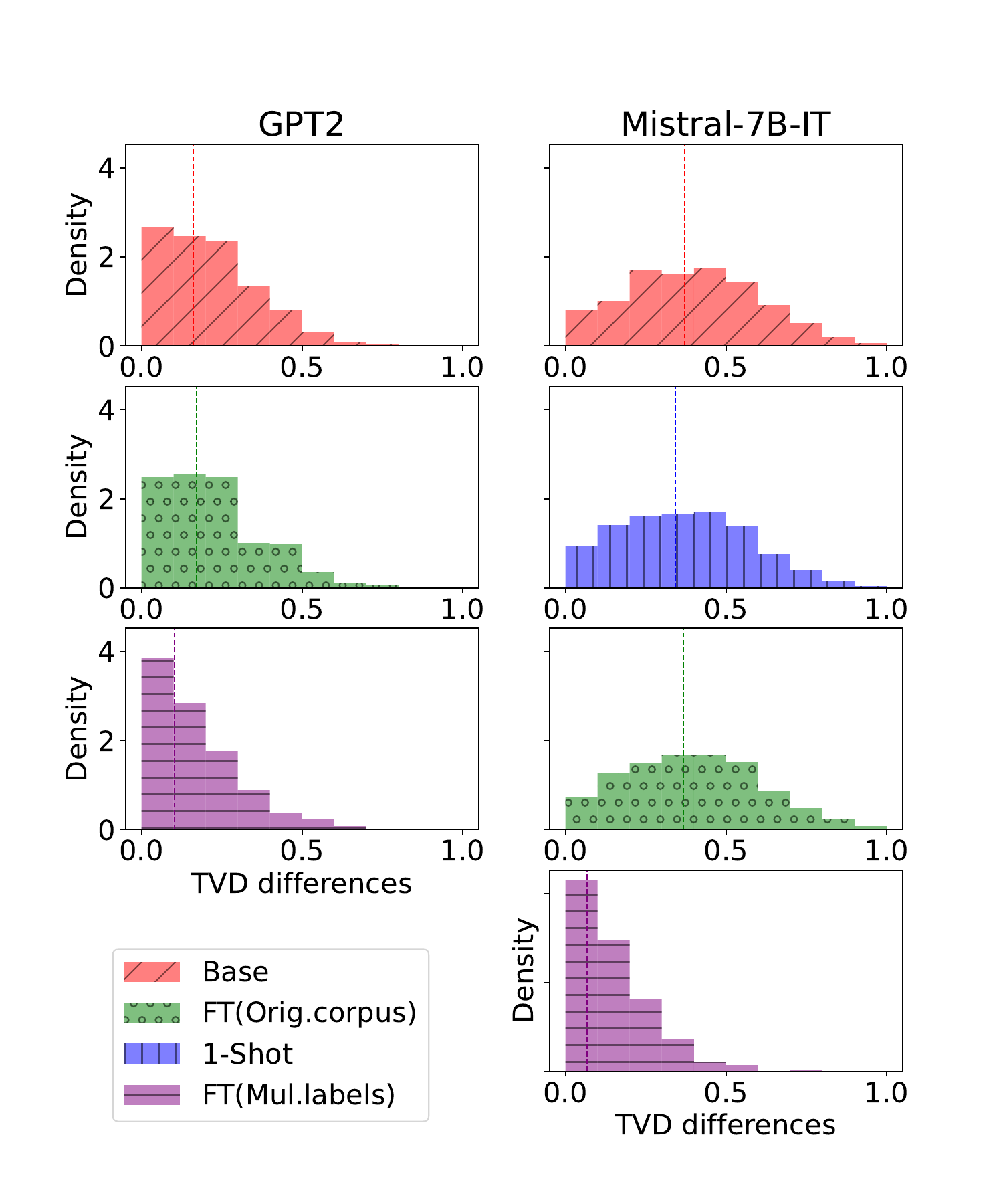}
  \end{subfigure}
  \hfill
  \centering
  \begin{subfigure}[t]{0.3\textwidth}
    \includegraphics[width=\linewidth]{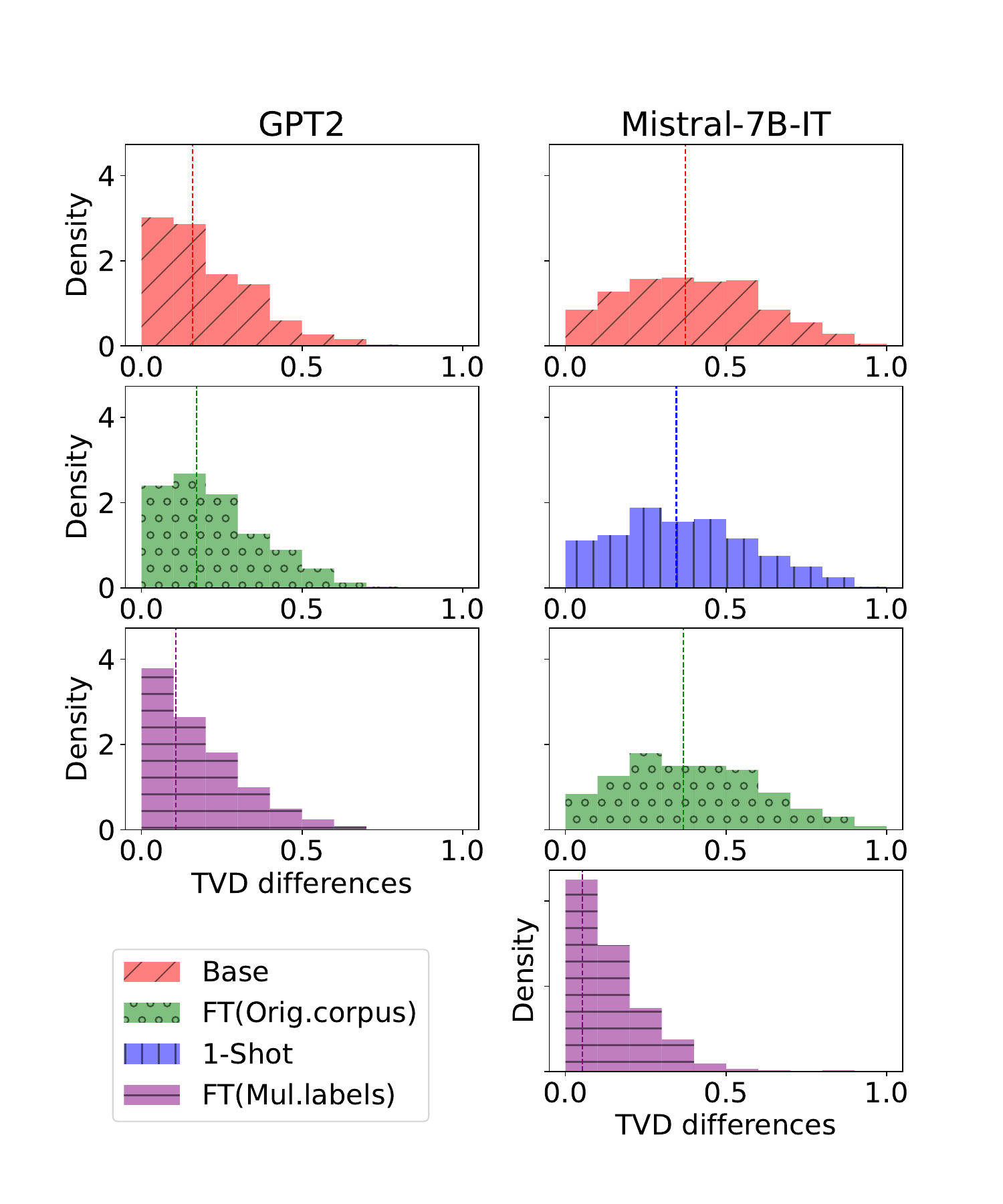}
  \end{subfigure}
  \hfill
  \begin{subfigure}[t]{0.3\textwidth}
    \includegraphics[width=\linewidth]{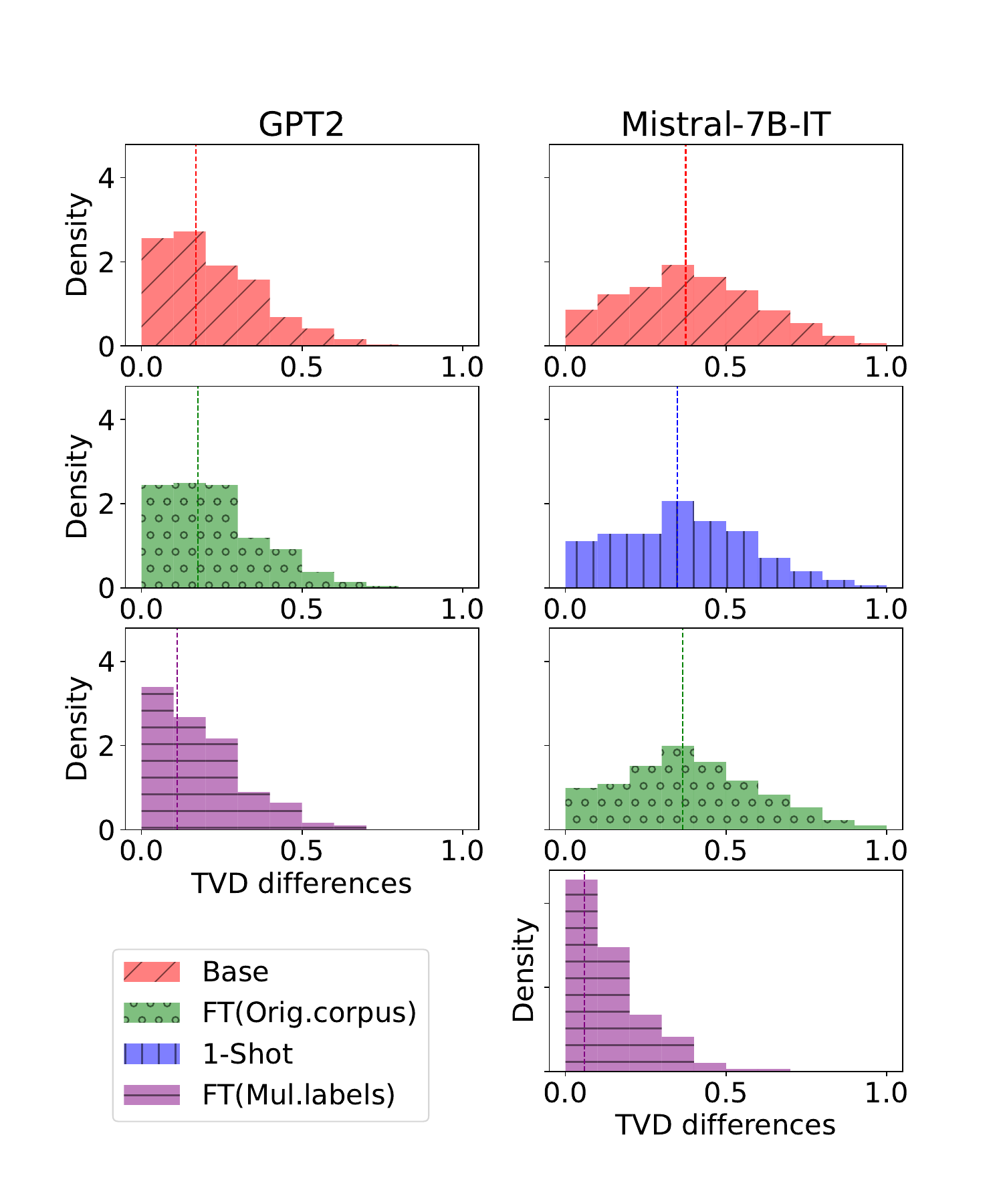}
    % \caption{mistral-7b-IT}
  \end{subfigure}
  \caption{Distribution of differences of TVD scores between the model and the human CPDs and the oracle CPDs, for all 3 seeds. For both GPT-2 and Mistral-7B-IT; fine-tuning shifts the TVD distribution towards smaller differences, confirming  previous findings.}
  \label{fig:differences}
\end{figure*}

We present \Cref{fig:tvd_main_combined}, which comprises the results on the test set for one of the three random seeds we trained on. We observe similar trends for the remaining seeds; which we present in \Cref{fig:tvd_main_seed2}. This is confirmed when plotting the differences between the TVD of the model and human CPDs and the TVD among the human oracle CPDs, as observed in \Cref{fig:differences}. 

\section{Analysis of model performance changes}
\label{app:anal_perform_changes}
\begin{figure*}[!ht]
  \centering
  \begin{subfigure}[t]{0.4\textwidth}
    \includegraphics[width=\linewidth]{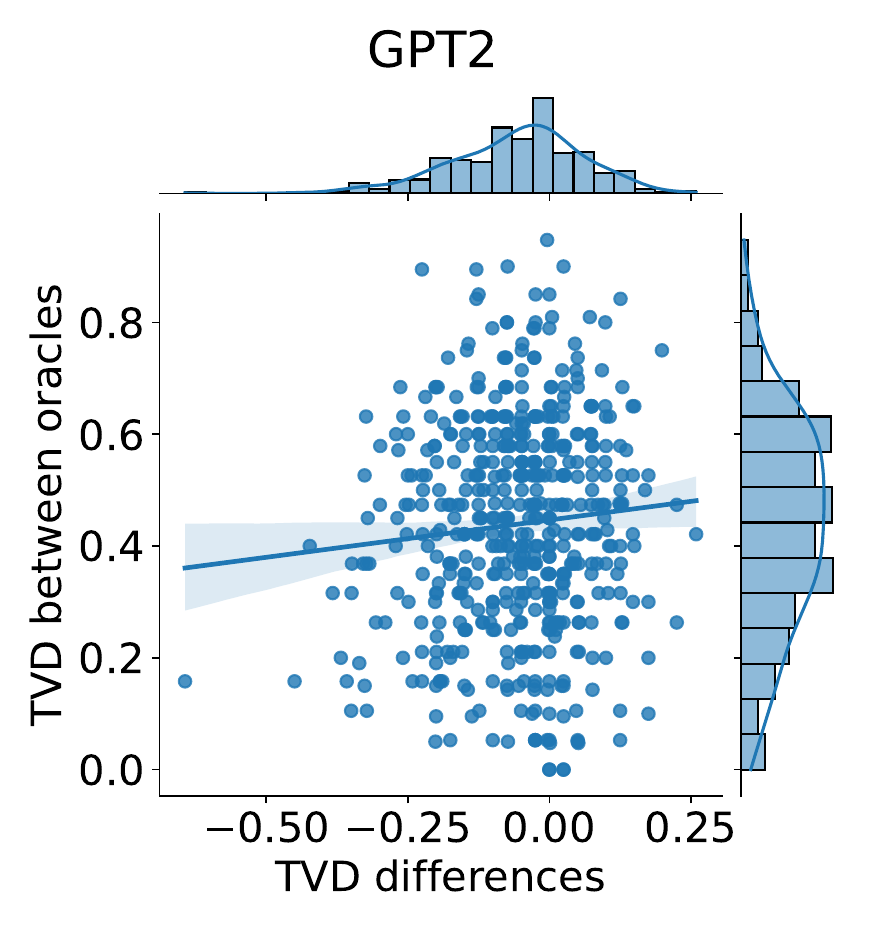}
  \end{subfigure}
  \hfill
  \centering
  \begin{subfigure}[t]{0.4\textwidth}
    \includegraphics[width=\linewidth]{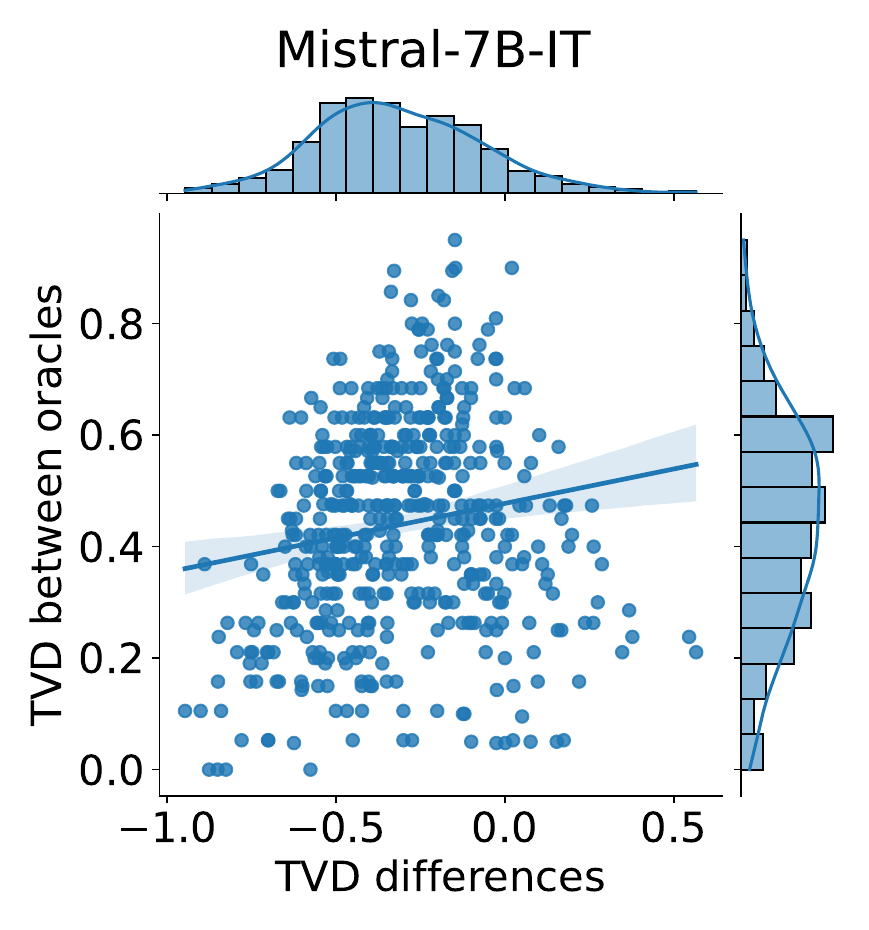}
  \end{subfigure}
  \hfill
  \centering
  \begin{subfigure}[t]{0.4\textwidth}
    \includegraphics[width=\linewidth]{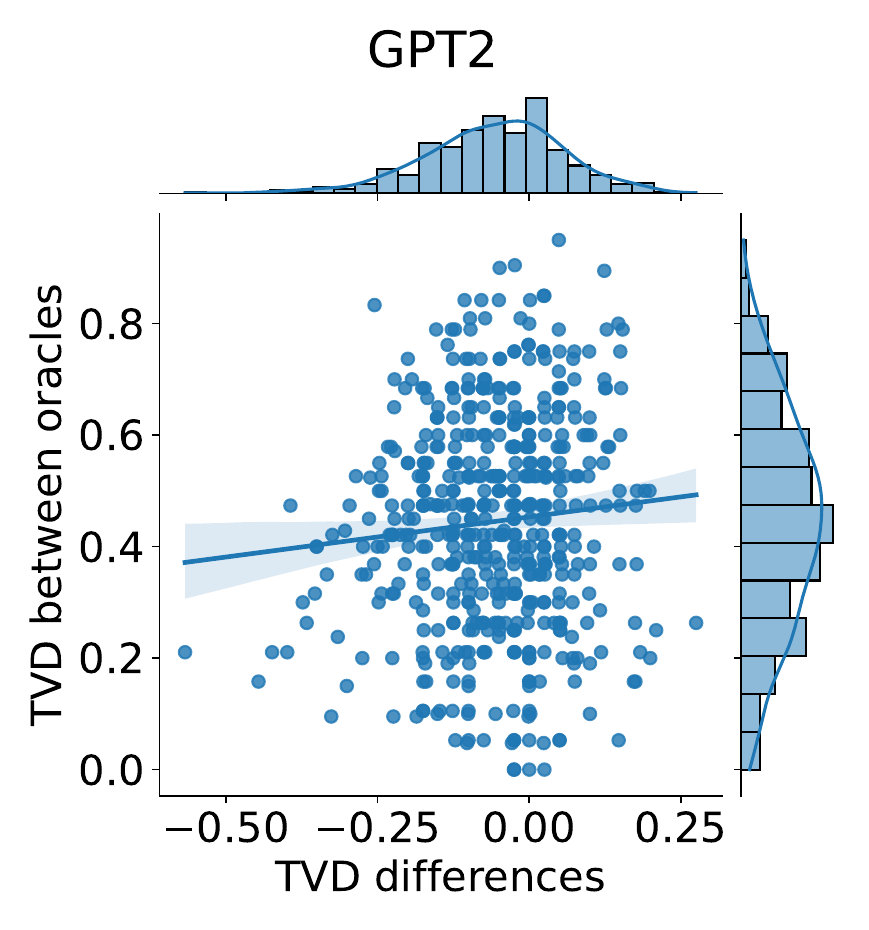}
  \end{subfigure}
  \hfill
  \centering
  \begin{subfigure}[t]{0.4\textwidth}
    \includegraphics[width=\linewidth]{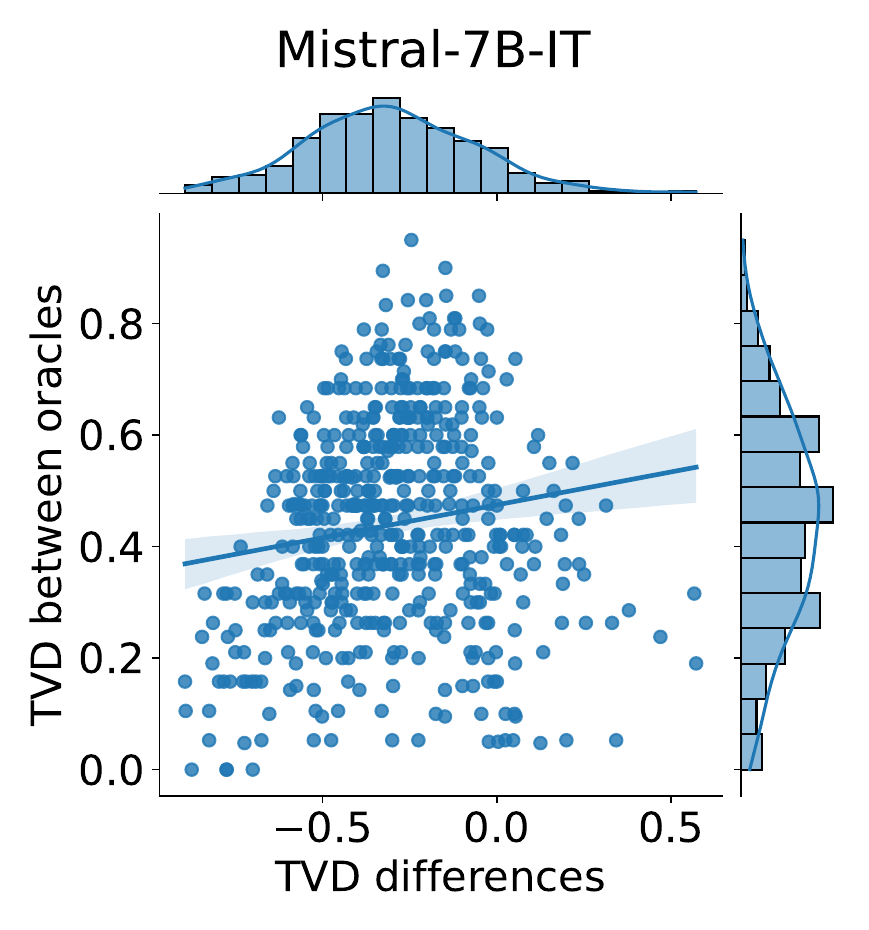}
  \end{subfigure}
  \hfill
  \centering
  \begin{subfigure}[t]{0.4\textwidth}
    \includegraphics[width=\linewidth]{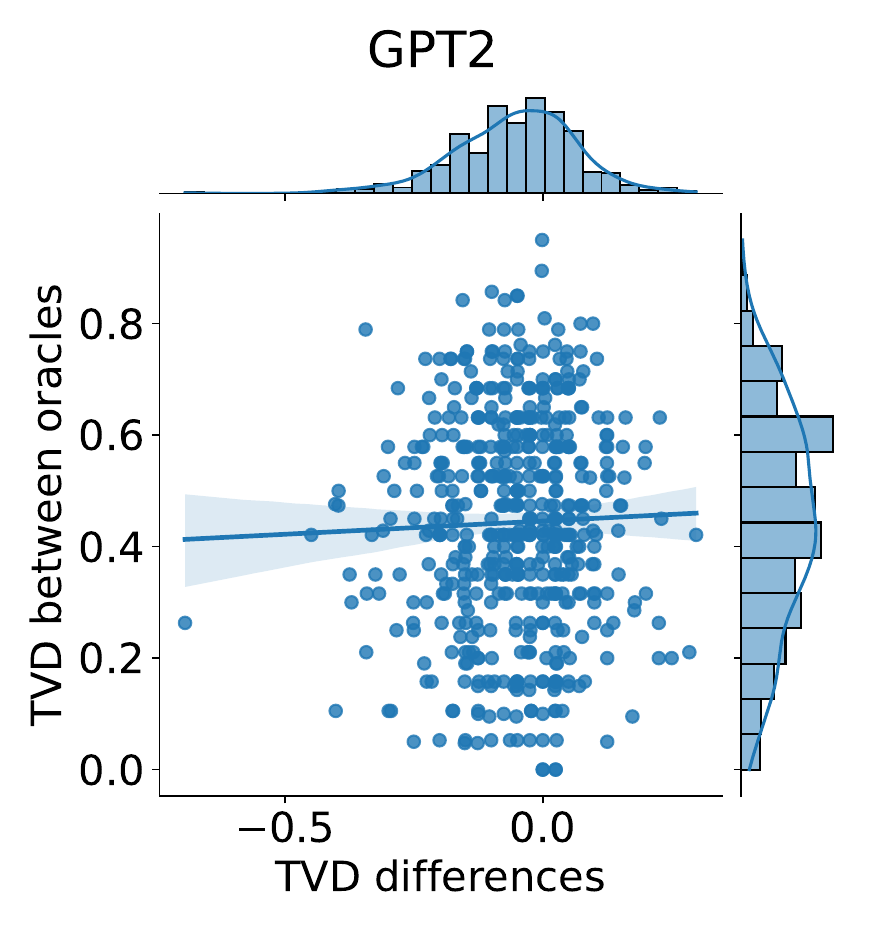}
  \end{subfigure}
  \hfill
  \centering
  \begin{subfigure}[t]{0.4\textwidth}
    \includegraphics[width=\linewidth]{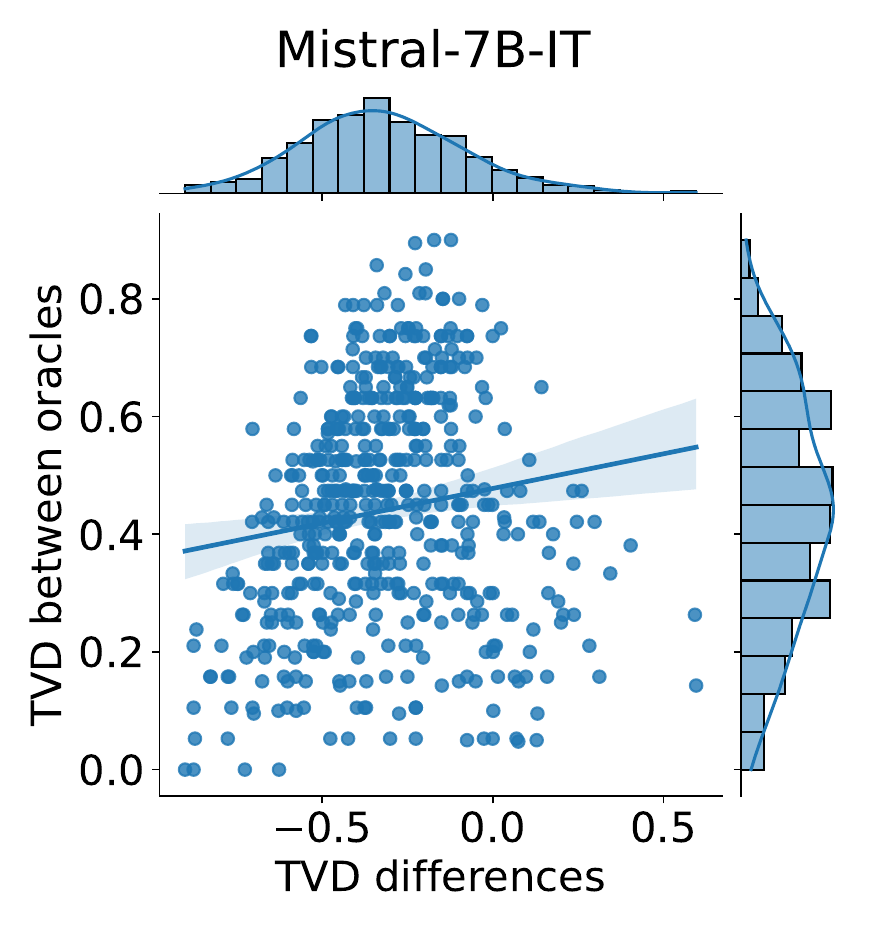}
  \end{subfigure}
  \hfill
  \caption{Distribution of differences of TVD scores between the fine tuned model and the human CPDs minus the TVD of the non fine tuned model and the human CPDs, against TVD among oracles. Performance gains (negative differences) for both models occur across contexts of varying open-endedness (with lower TVD indicating more `restricted' contexts).}
  \label{fig:differences_tvd_vs_tvdoracles}
\end{figure*}

\begin{figure*}[!ht]
  \centering
  \begin{subfigure}[t]{0.4\textwidth}
    \includegraphics[width=\linewidth]{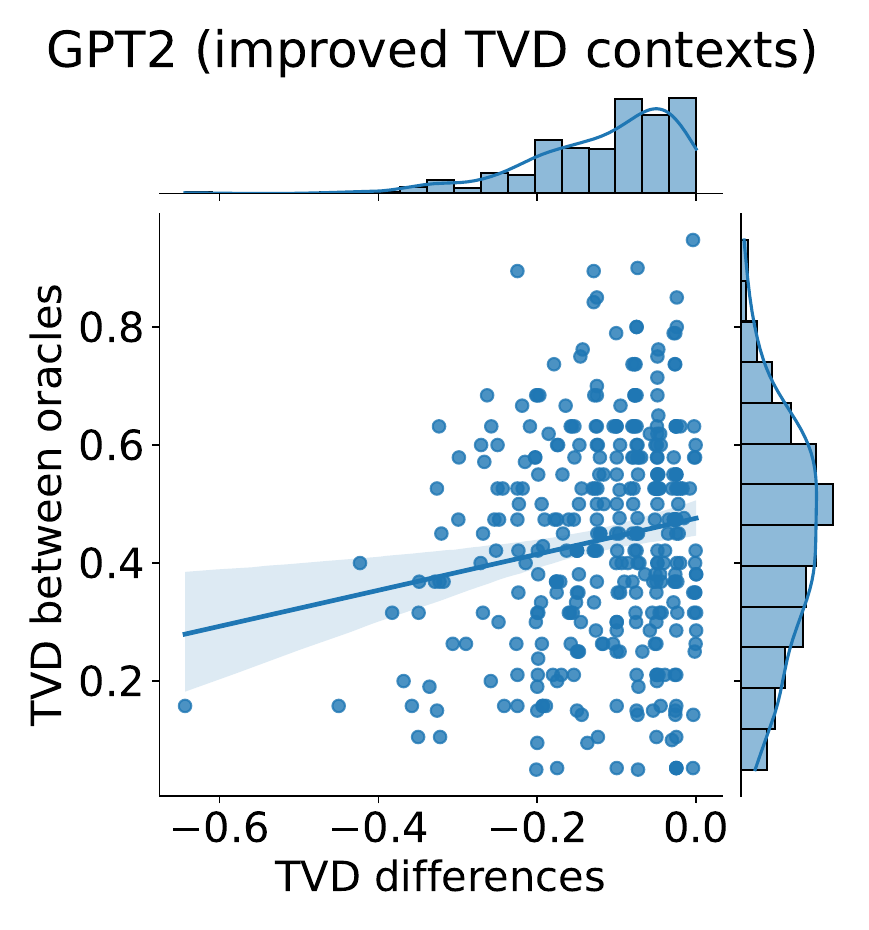}
  \end{subfigure}
  \hfill
  \centering
  \begin{subfigure}[t]{0.45\textwidth}
    \includegraphics[width=\linewidth]{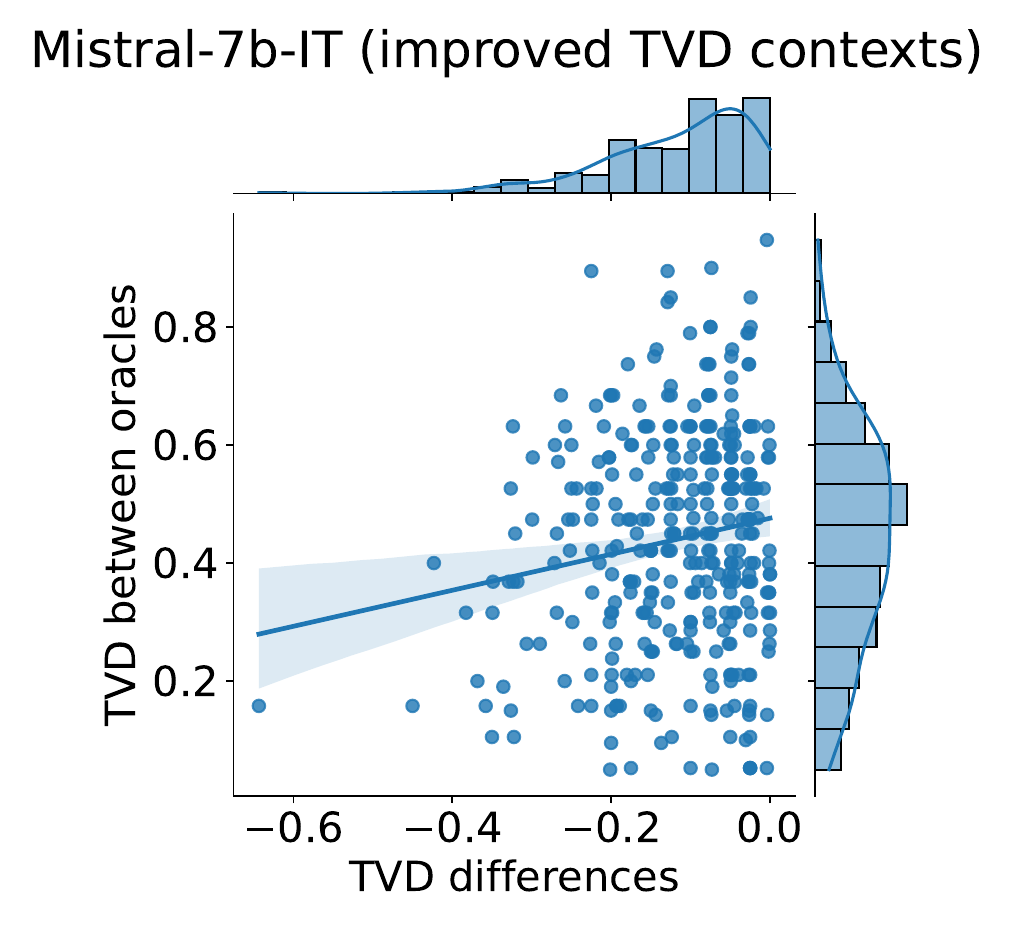}
  \end{subfigure}
  \hfill
    \centering
  \begin{subfigure}[t]{0.4\textwidth}
    \includegraphics[width=\linewidth]{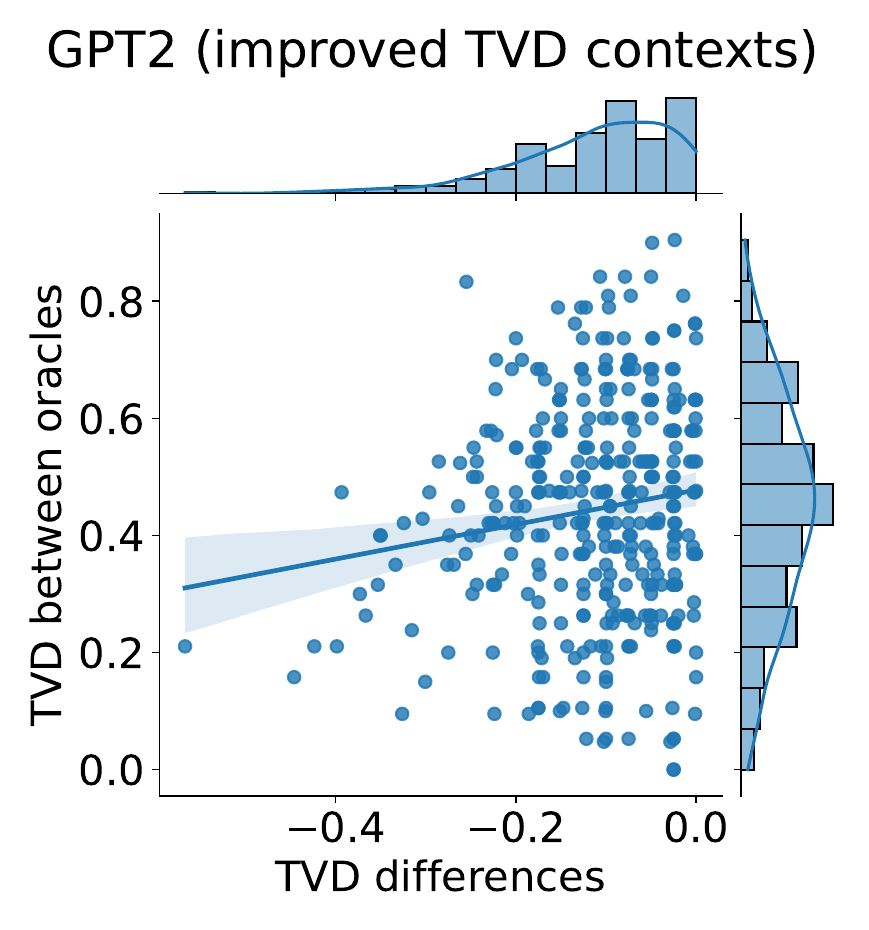}
  \end{subfigure}
  \hfill
  \centering
  \begin{subfigure}[t]{0.45\textwidth}
    \includegraphics[width=\linewidth]{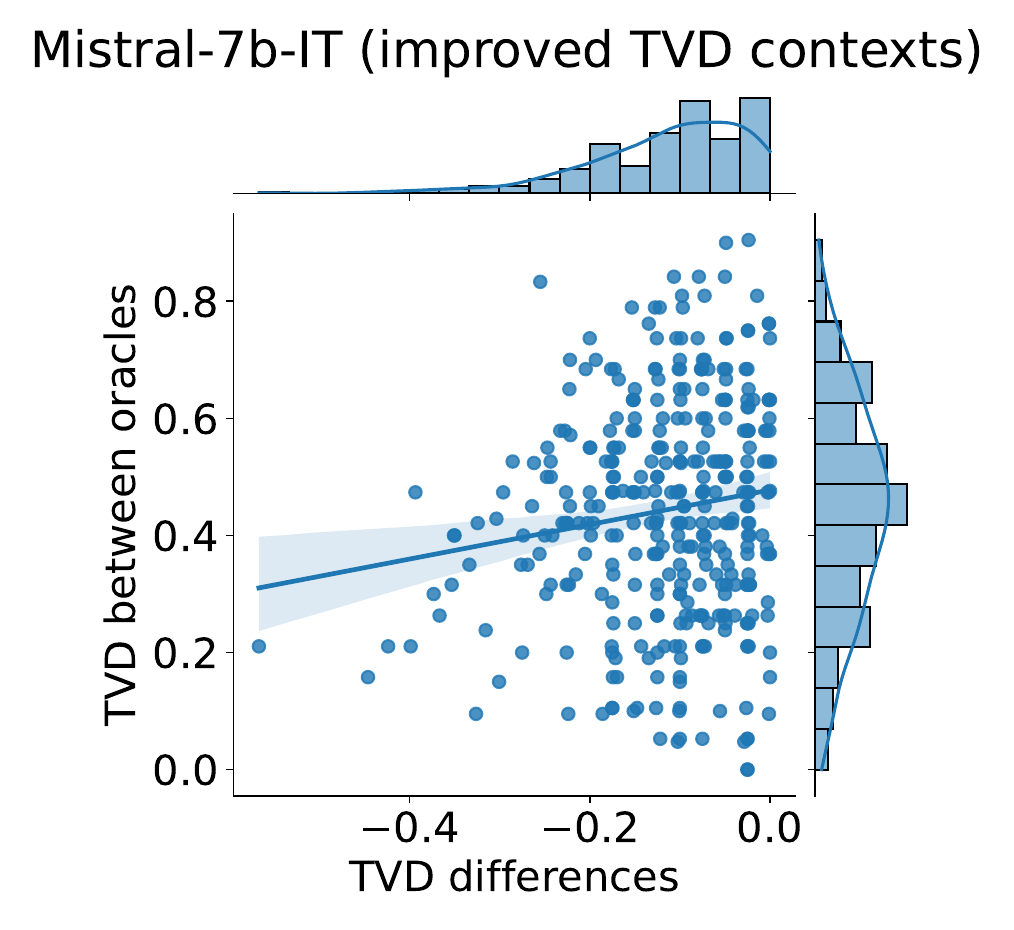}
  \end{subfigure}
  \hfill
    \centering
  \begin{subfigure}[t]{0.4\textwidth}
    \includegraphics[width=\linewidth]{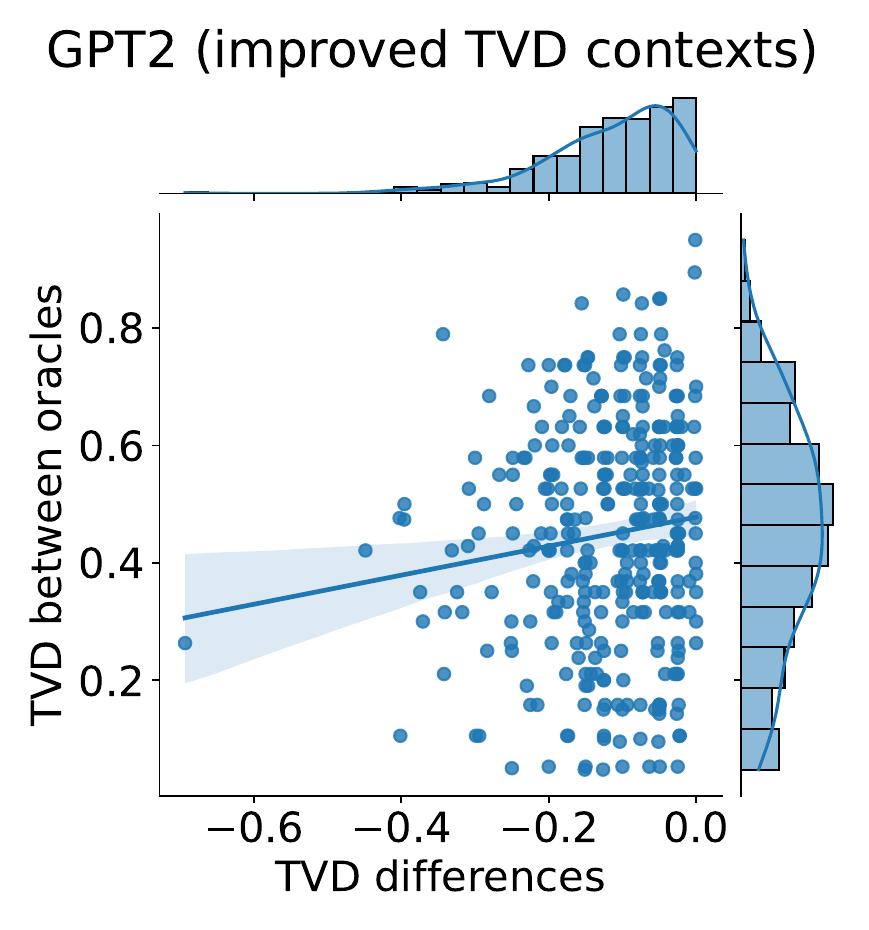}
  \end{subfigure}
  \hfill
  \centering
  \begin{subfigure}[t]{0.45\textwidth}
    \includegraphics[width=\linewidth]{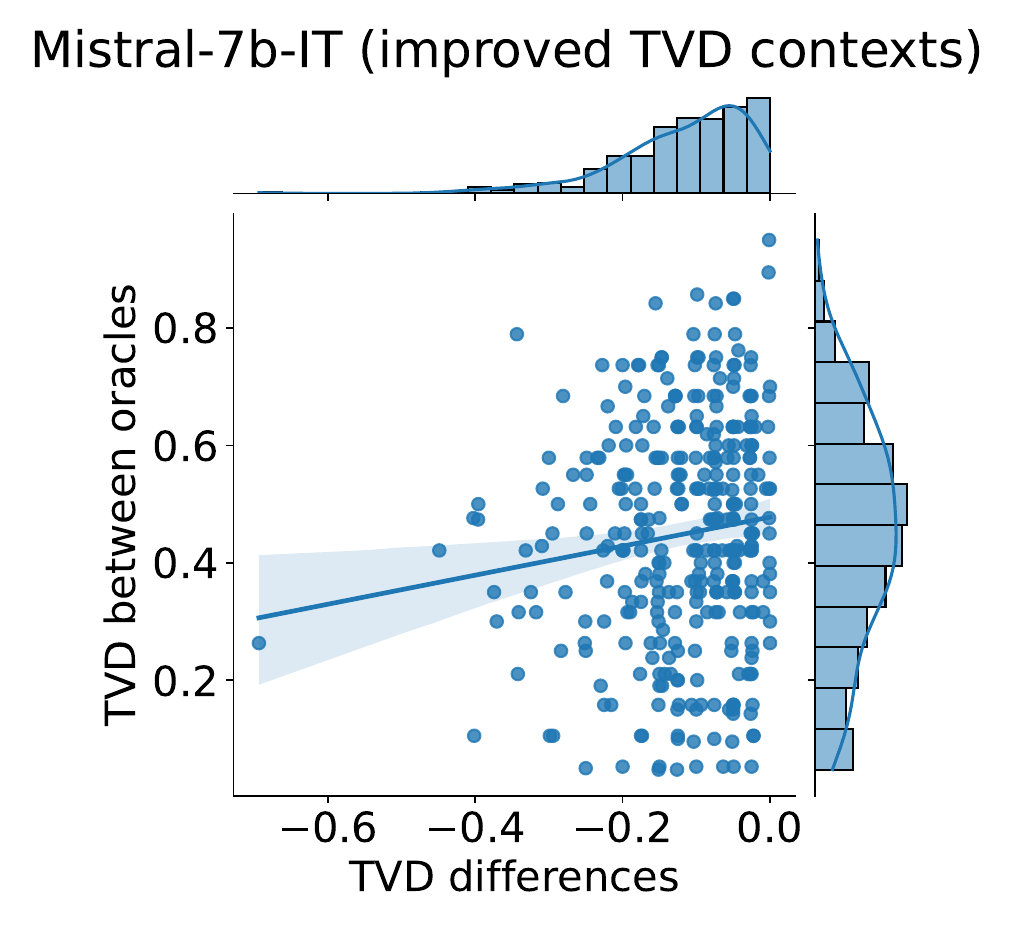}
  \end{subfigure}
  \hfill
  \caption{Distribution of differences of TVD scores between the fine tuned model and the human CPDs minus the TVD of the non fine tuned model and the human CPDs, against TVD among oracles. In this case, we only plot datapoints for which we observed improvements (\emph{i.e.} negative differences) for both models. Similarly, we observe that gains occur across contexts of varying open-endedness (with lower TVD indicating more `restricted' contexts).}
  \label{fig:gains_tvd_vs_tvdoracles}
\end{figure*}

In order to understand how fine tuning has affected the model performance. We perform various analyses.  
We visualise the models' changes in performance against context open-endedness. We approximate that using the TVD between human oracles. We assume that a lower TVD, reflecting lower disagreement among human populations, indicates more `restrictive' contexts, while a higher TVD, indicates contexts that admit a higher level of plausible variability. 
We plot changes in performance by computing the differences between the TVD of the fine tuned model and human CPD and the TVD of the non fine tuned model and human CPD. Results are shown in \Cref{fig:differences_tvd_vs_tvdoracles} (showing all contexts) and \Cref{fig:gains_tvd_vs_tvdoracles} (showing only contexts for which performance improved, \emph{i.e.} negative differences in TVD values). We observe how improvements occur across contexts of varying open-endedness (\emph{i.e.} varying TVD among oracles values). 

\begin{figure*}[t]
  \centering
  \begin{subfigure}[t]{0.48\textwidth}
    \includegraphics[width=\linewidth]{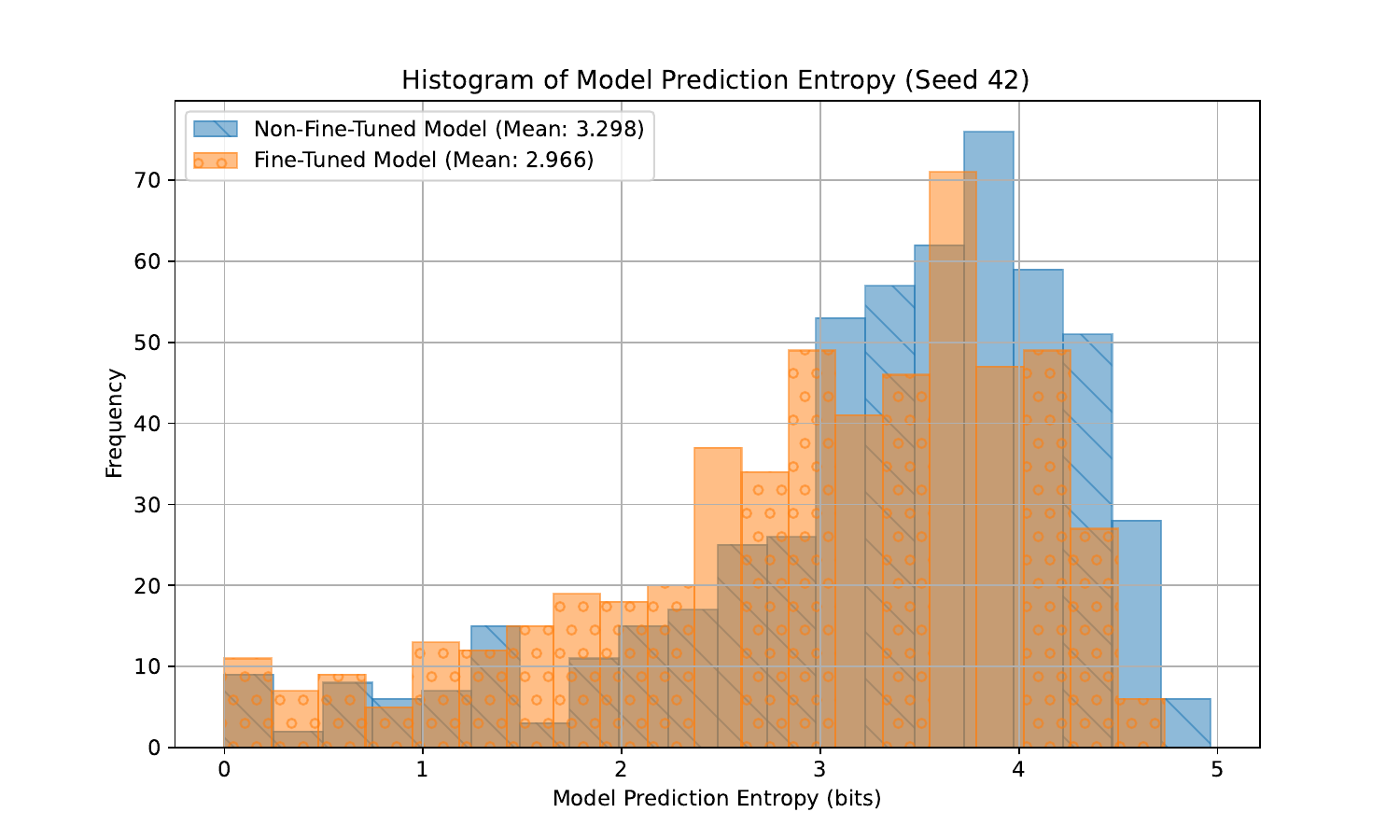}
    \caption{GPT-2 (Seed 42)}
  \end{subfigure}
  \hfill
  \begin{subfigure}[t]{0.45\textwidth}
    \includegraphics[width=\linewidth]{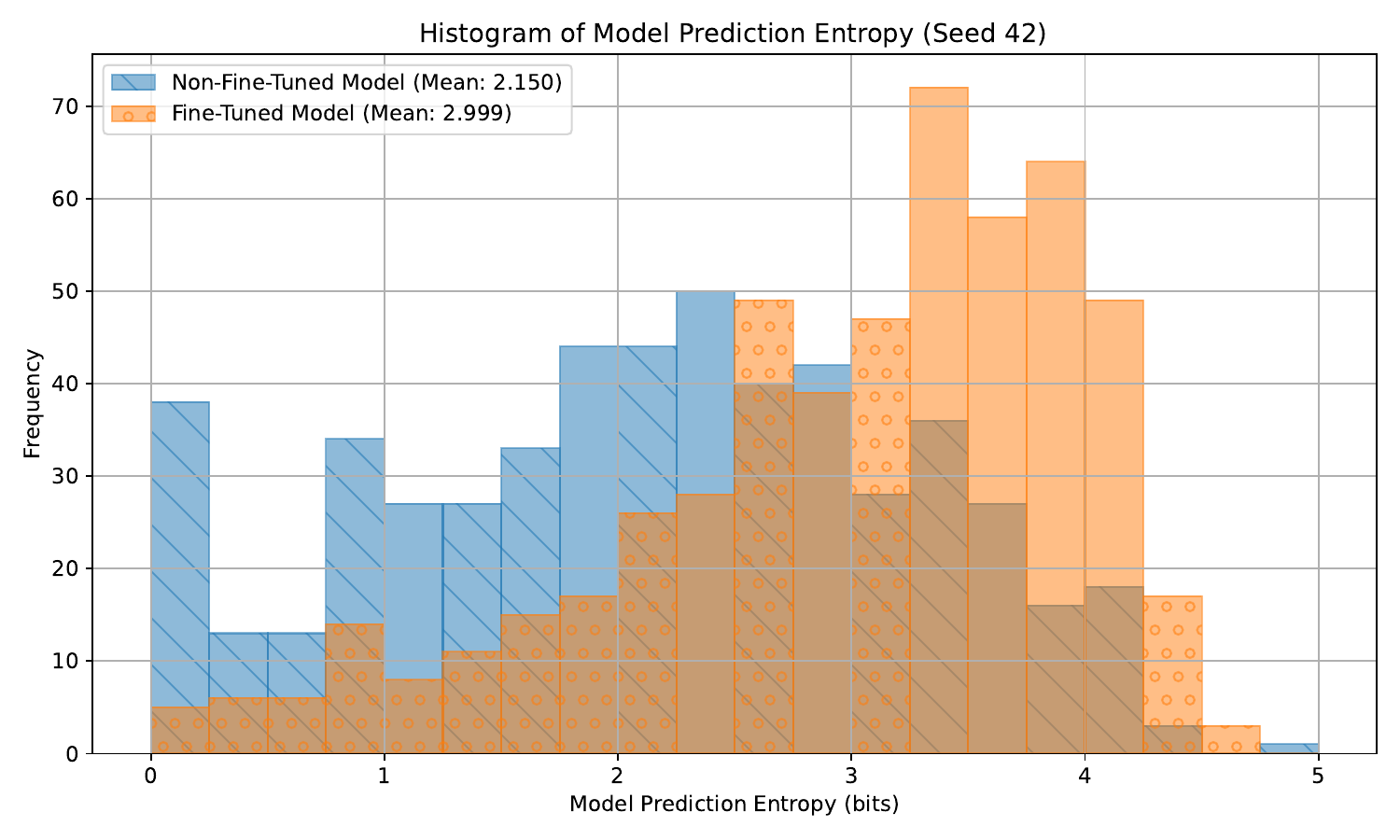}
    \caption{Mistral-7B (Seed 42)}
  \end{subfigure}
  \vspace{1em} % Adds vertical space between rows
  \begin{subfigure}[t]{0.48\textwidth}
    \includegraphics[width=\linewidth]{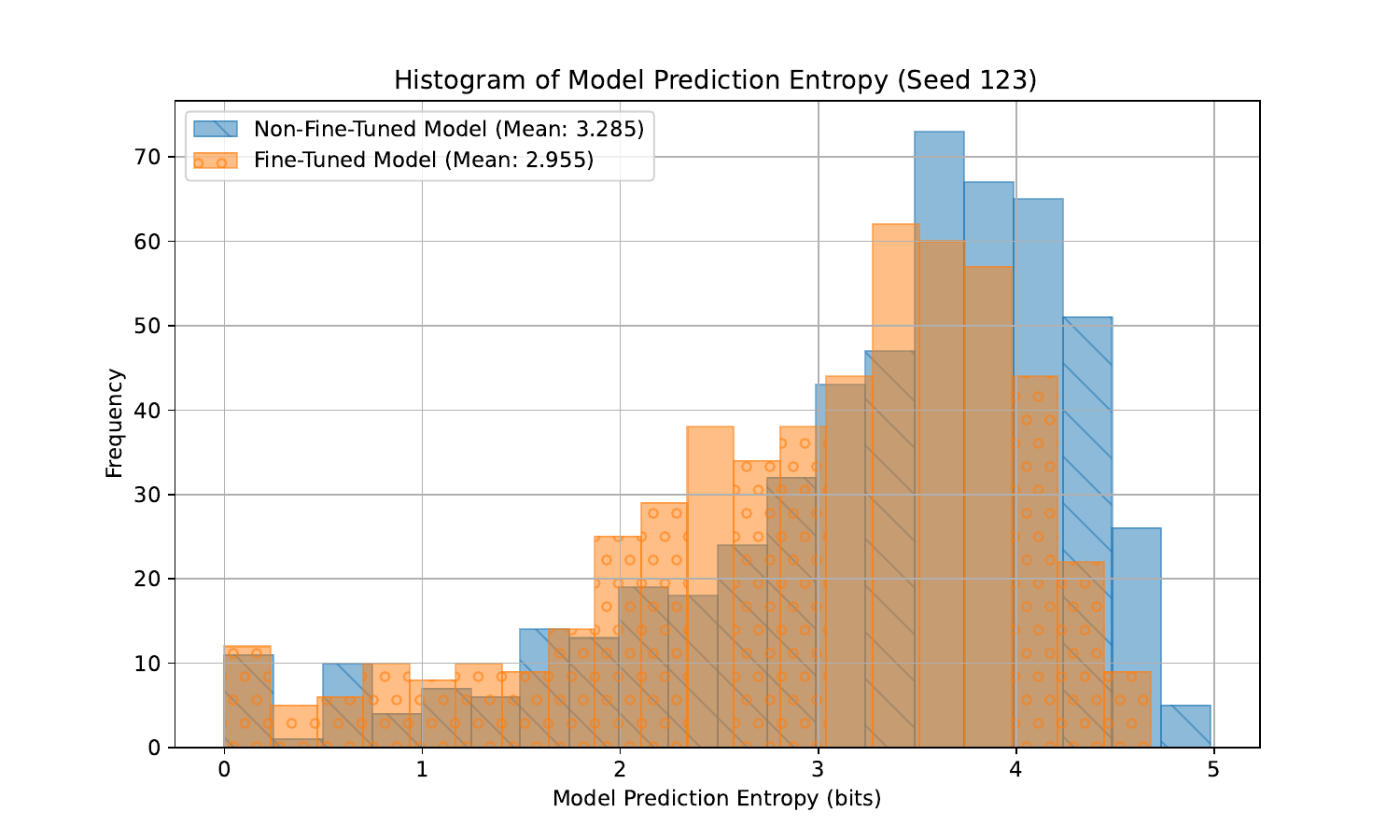}
    \caption{GPT-2 (Seed 123)}
  \end{subfigure}
  \hfill
  \begin{subfigure}[t]{0.45\textwidth}
    \includegraphics[width=\linewidth]{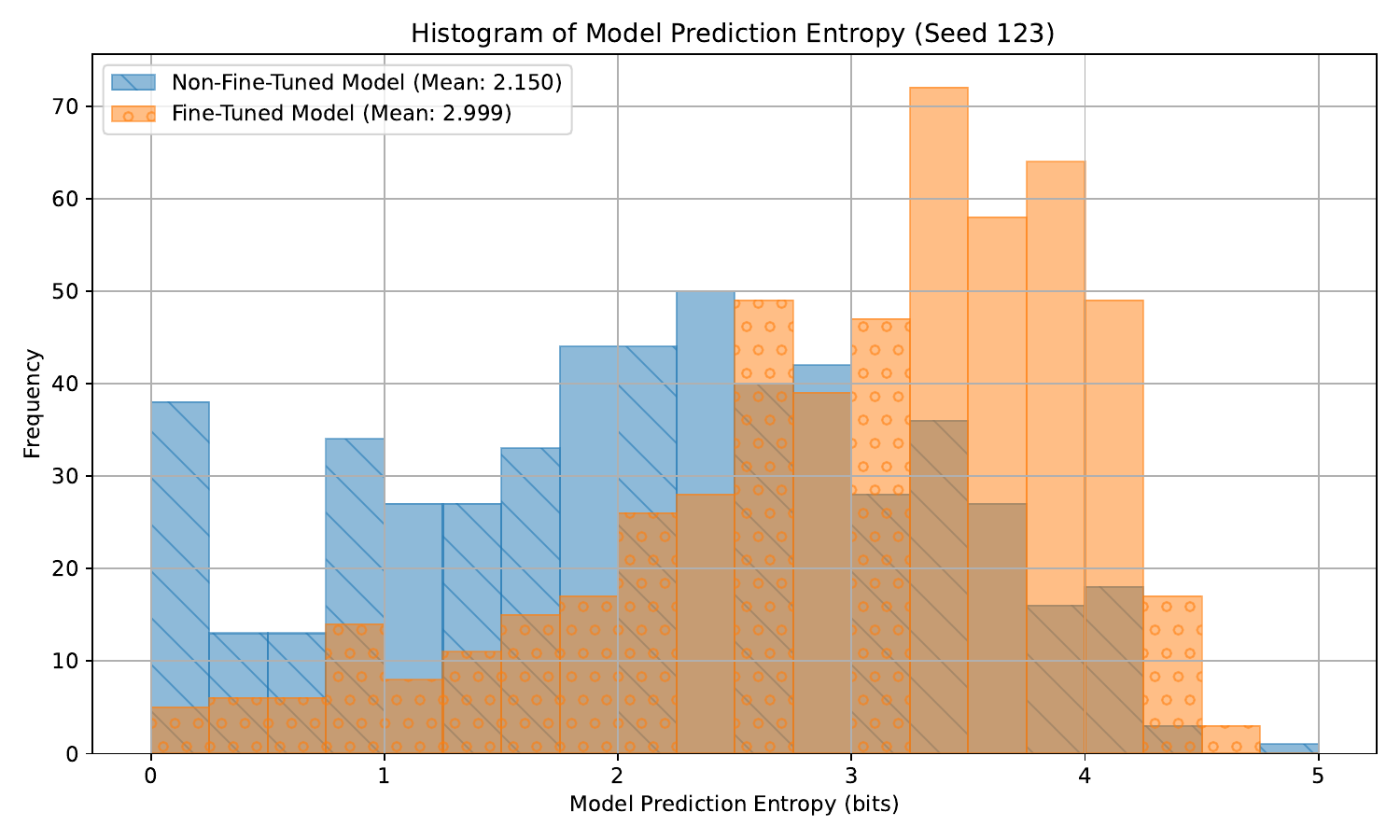}
    \caption{Mistral-7B (Seed 123)}
  \end{subfigure}
  \vspace{1em} % Adds vertical space between rows
  \begin{subfigure}[t]{0.48\textwidth}
    \includegraphics[width=\linewidth]{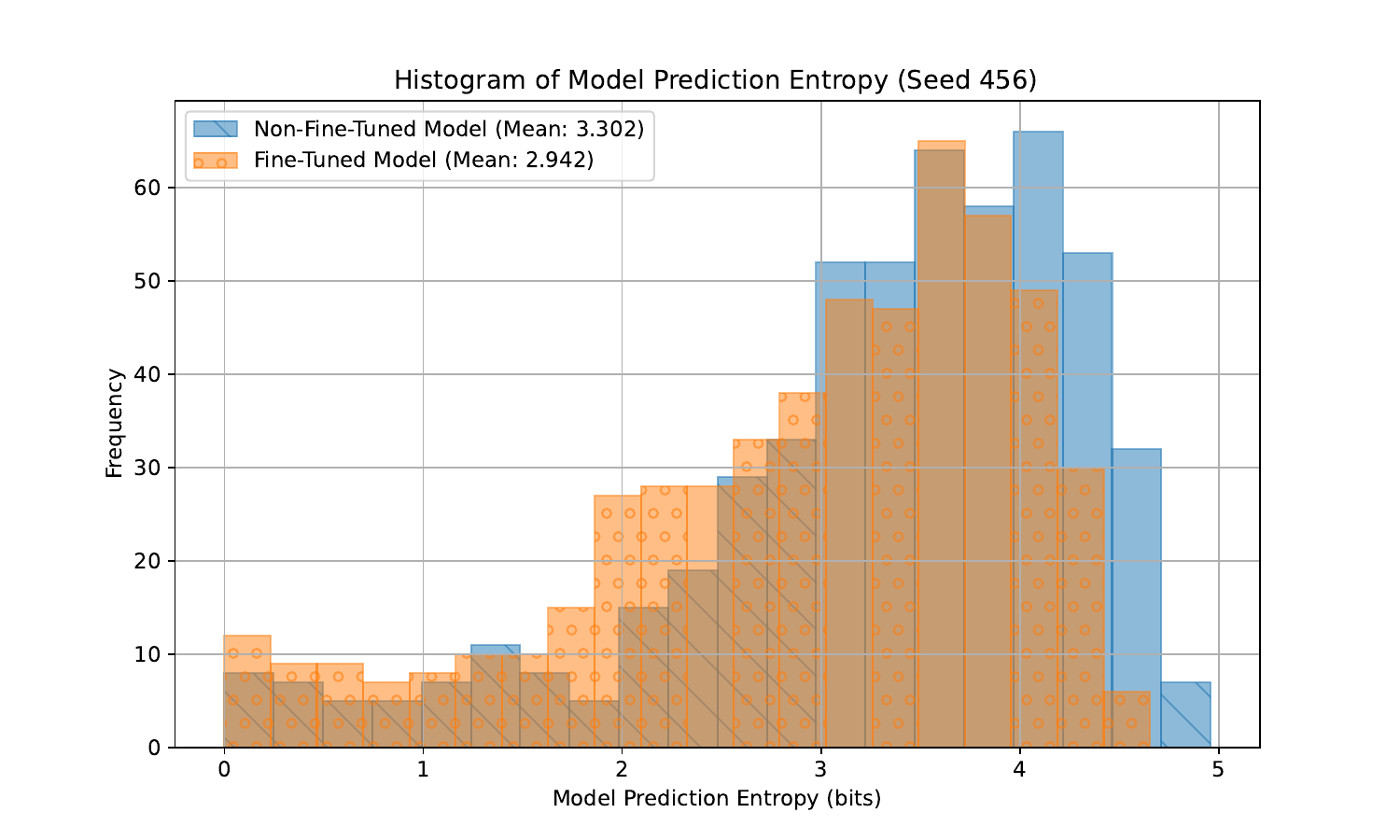}
    \caption{GPT-2 (Seed 456)}
  \end{subfigure}
  \hfill
  \begin{subfigure}[t]{0.45\textwidth}
    \includegraphics[width=\linewidth]{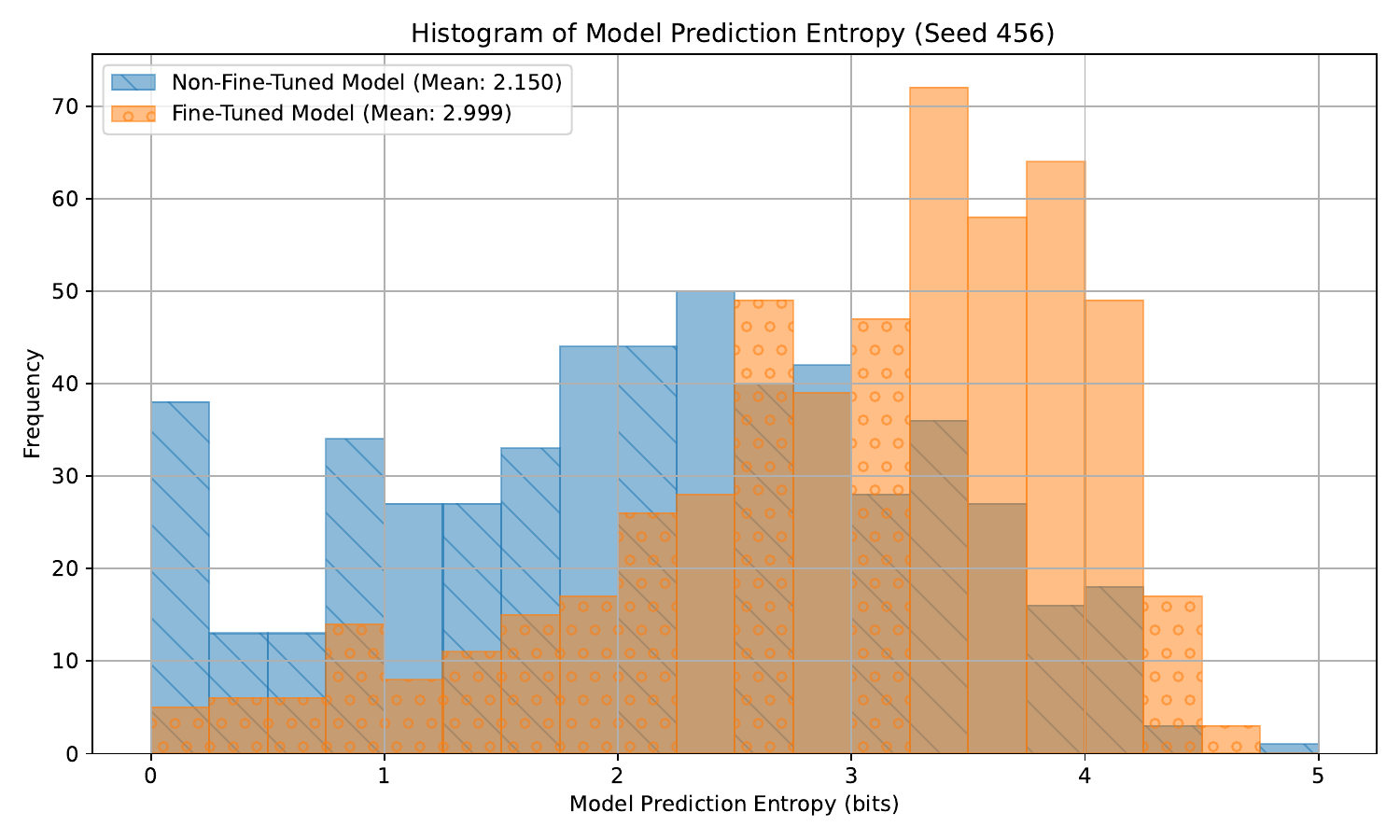}
    \caption{Mistral-7B (Seed 456)}
  \end{subfigure}
  \caption{Entropy of model predictions before an after finetuning.}
  \label{fig:model_entropy_hist}
\end{figure*}

To gain further insight as to how fine tuning has affected our models, we plot the entropy values of the empirically estimated model CPDs across contexts before and after fine-tuning. Results can be seen in Figure~\ref{fig:model_entropy_hist}. For GPT2, we observe how the entorpy of the model's empirically estimated CPDs were not impacted very substantially. We observe only a slight shift towards lower entropy values (\emph{i.e.} peakier distributions); which means that model predictions might be slightly more confident, while also being better better aligned with human linguistic variability. On the contrary, the fine tuned Mistral-7B-IT model's entropy values shift substantially towards higher values, demonstrating that now the model is making more diverse predictions (which are also better aligned with human linguistic variability, as evident by our main findings). 

%mistral is making very confident predictions (i.e. low entropy) before -> afterwords more diverse distributions

\begin{figure*}[t]
  \centering
  \begin{subfigure}[t]{0.48\textwidth}
    \includegraphics[width=\linewidth]{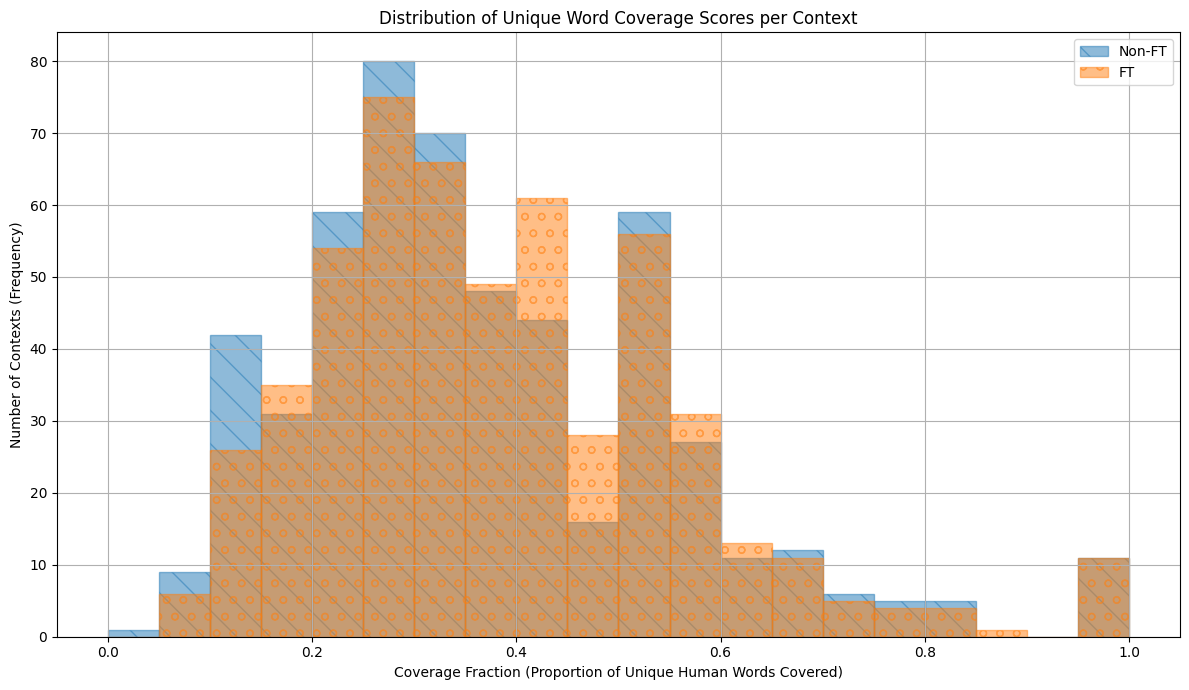}
    \caption{GPT-2}
    \label{fig:unique_dominant_word_gpt2}
  \end{subfigure}
  \hfill
  \begin{subfigure}[t]{0.48\textwidth}
    \includegraphics[width=\linewidth]{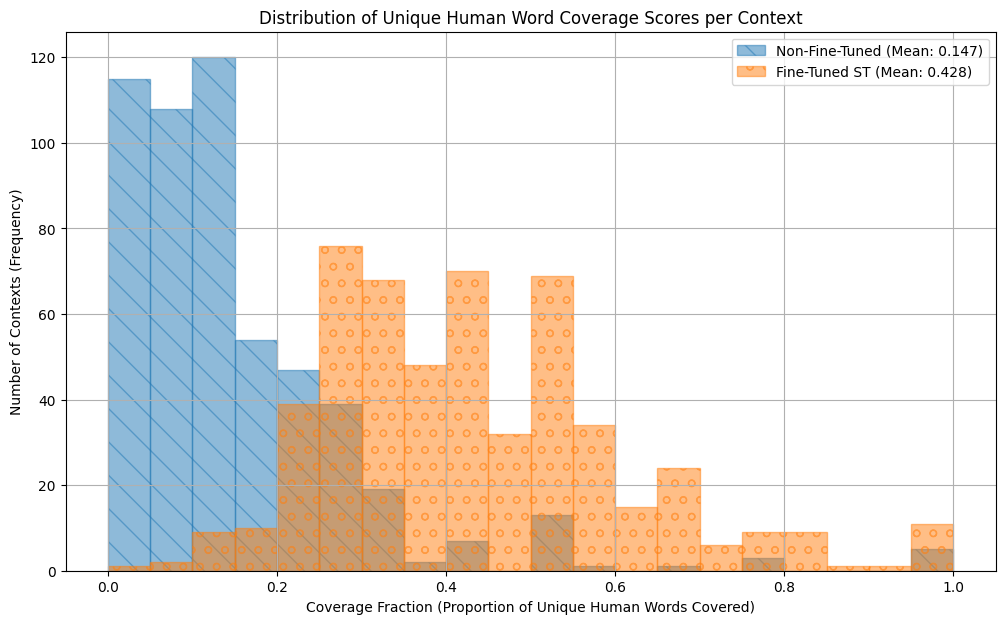}
    \caption{Mistral-7B}
    \label{fig:unique_word_hist_mistral}
  \end{subfigure}
  \caption{Unique word coverage across models. Fine-tuning with multiple labels per instance increases lexical diversity more compared to hard-targets (majority vote).}
  \label{fig:unique_word_hist_combined}
\end{figure*}

Lastly, we assess whether models improve at predicting words that humans predicted (regardless of their frequency), as a means to approximate how wel their lexical diversity aligns with that of our assessed human population. We plot the fraction of unique human predictions that were also predicted by the models before and after fine-tuning with multiple labels (\Cref{fig:unique_word_hist_combined}). Higher values indicating a more highly aligned lexical diversity.
We find that GPT2's lexical diversity remained relatively similar to before fine tuning, but for Mistral-7B-IT we see a clear rightward shift in the distribution of unique word coverage for the fine-tuned model. This indicates that the fine-tuned model predicts a greater number of relevant unique words per context compared to the non-fine-tuned baseline. 

\section{Analysis of model fine-tuned with majority label}
\label{App:majlabelFT}

\begin{figure*}[!ht]
  \centering
  \begin{subfigure}[t]{0.4\textwidth}
    \includegraphics[width=\linewidth]{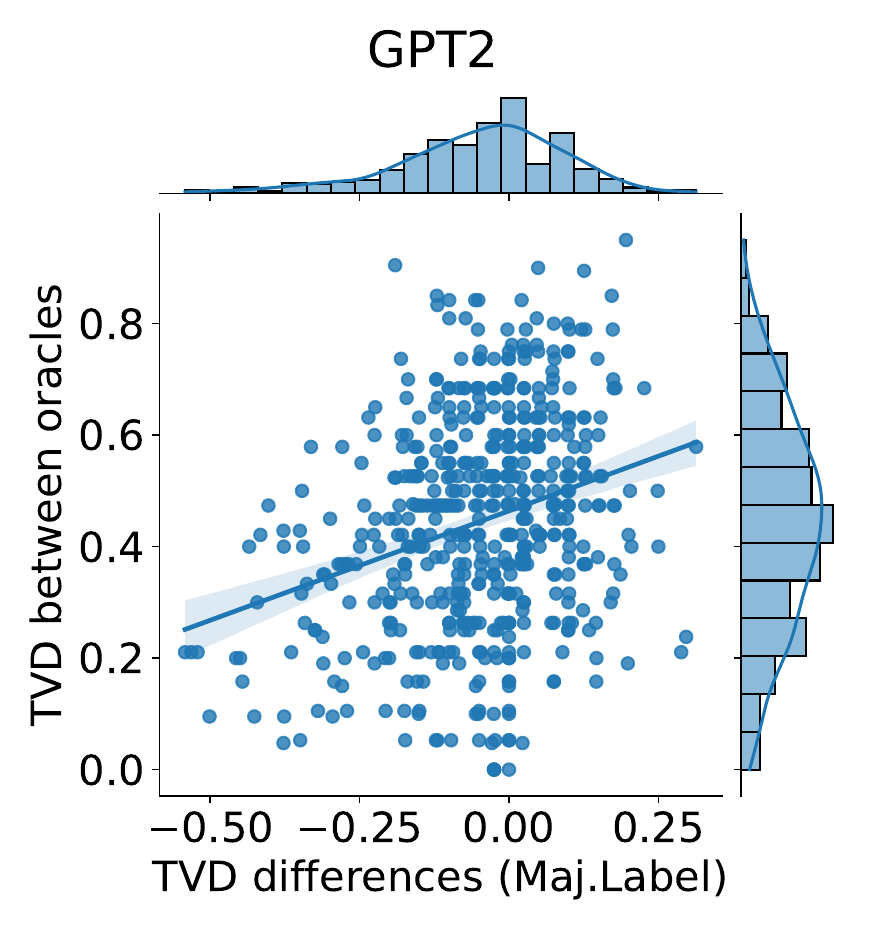}
  \end{subfigure}
  \hfill
  \centering
  \begin{subfigure}[t]{0.4\textwidth}
    \includegraphics[width=\linewidth]{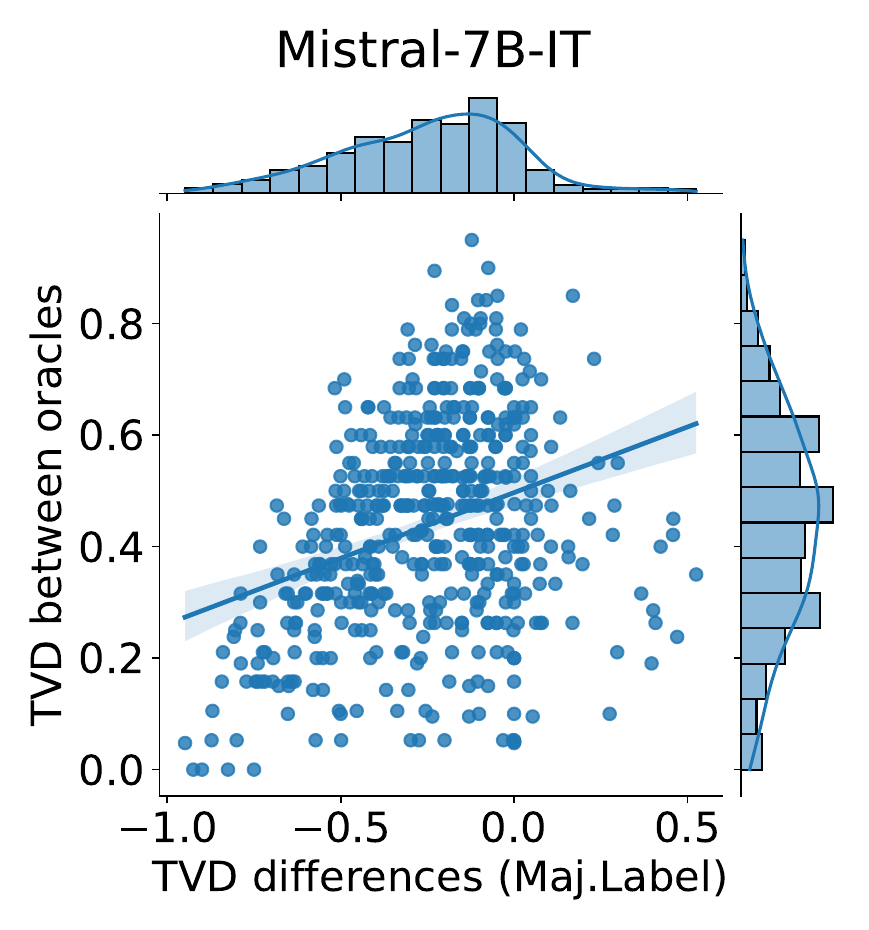}
  \end{subfigure}
  \hfill
  \centering
  \begin{subfigure}[t]{0.4\textwidth}
    \includegraphics[width=\linewidth]{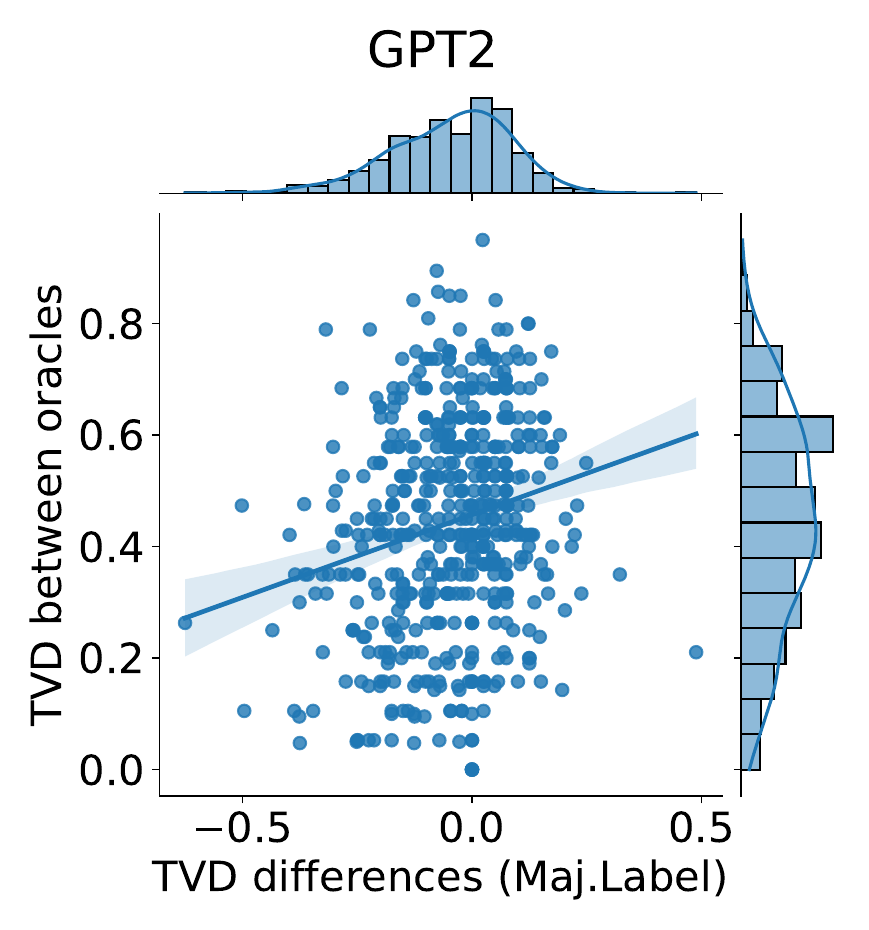}
  \end{subfigure}
  \hfill
  \centering
  \begin{subfigure}[t]{0.4\textwidth}
    \includegraphics[width=\linewidth]{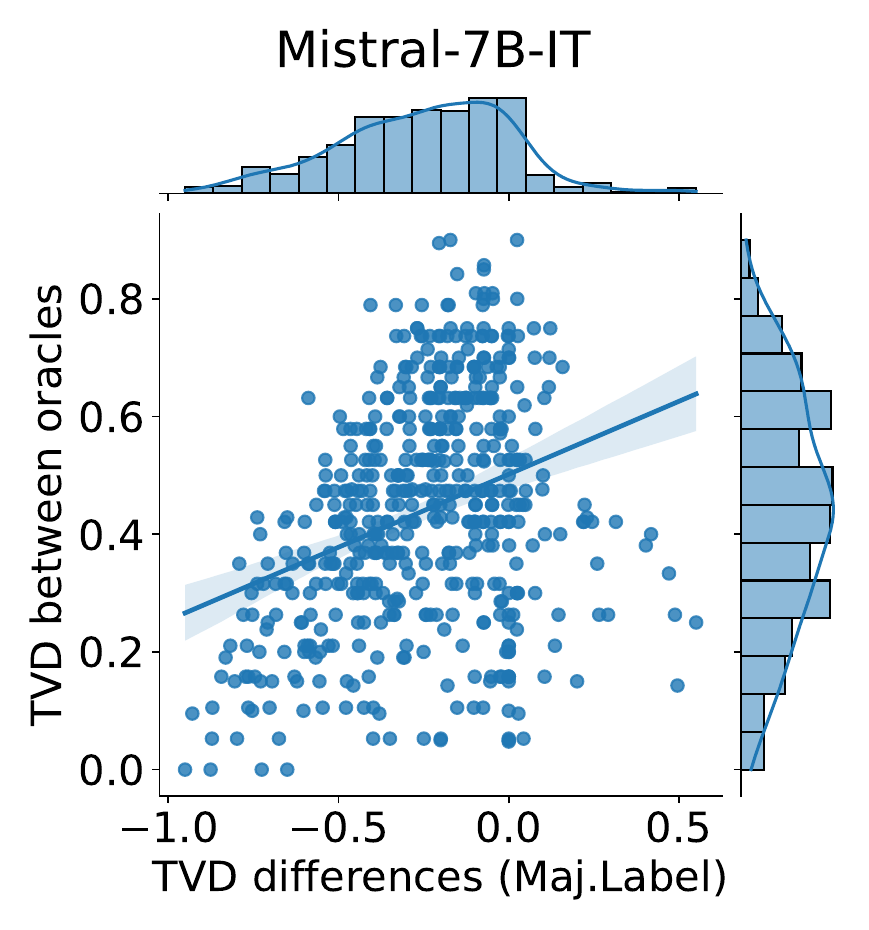}
  \end{subfigure}
  \hfill
  \centering
  \begin{subfigure}[t]{0.4\textwidth}
    \includegraphics[width=\linewidth]{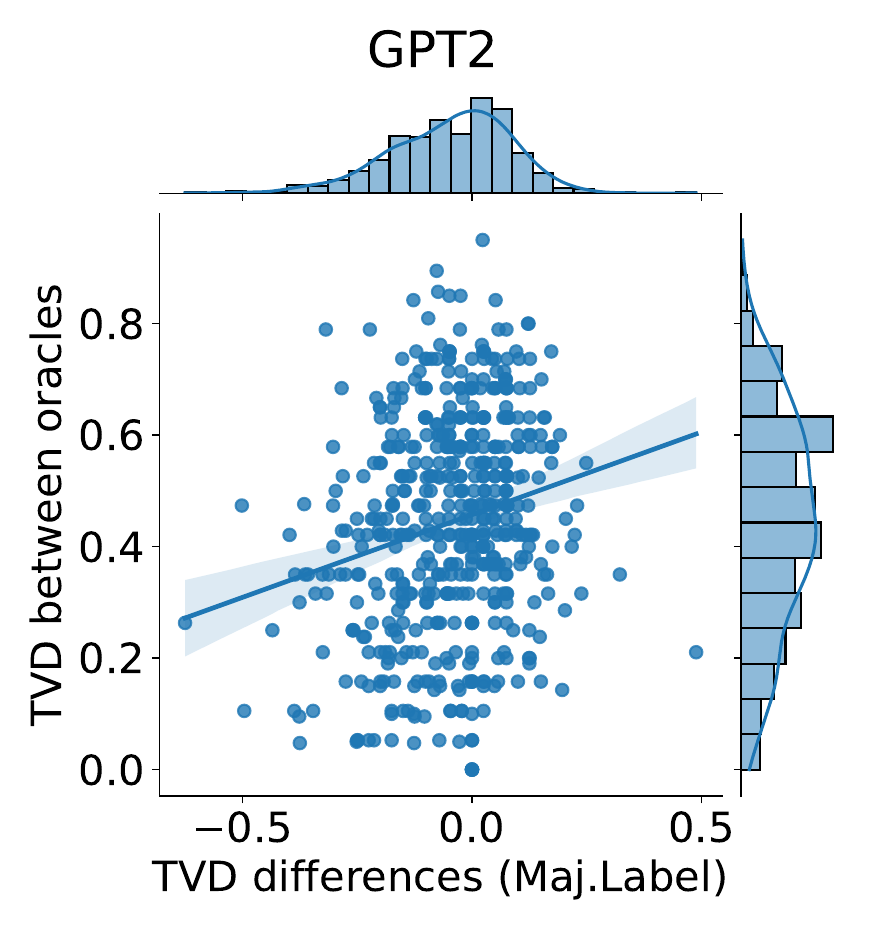}
  \end{subfigure}
  \hfill
  \centering
  \begin{subfigure}[t]{0.4\textwidth}
    \includegraphics[width=\linewidth]{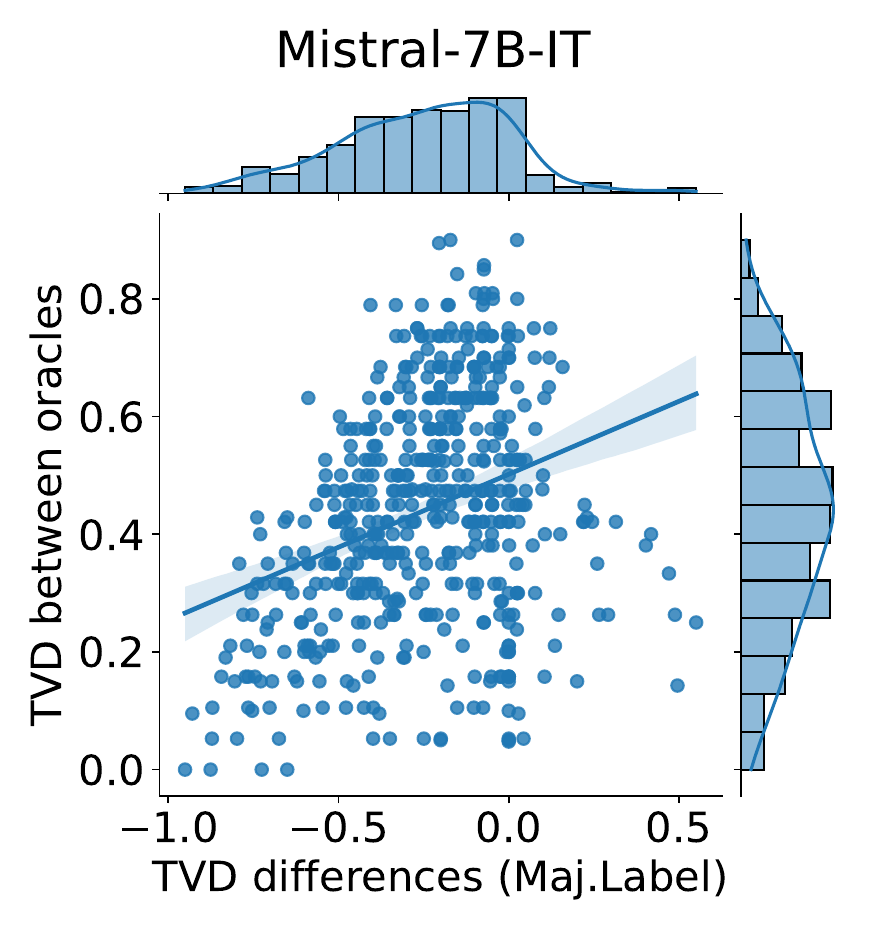}
  \end{subfigure}
  \hfill
  \caption{Distribution of differences of TVD scores between the fine tuned model and the human CPDs minus the TVD of the non fine tuned model and the human CPDs, against TVD among oracles. Performance gains (negative differences) for both models occur across contexts of varying open-endedness (with lower TVD indicating more `restricted' contexts).}
  \label{fig:differences_tvd_vs_tvdoracles_majlabel}
\end{figure*}

Similar to Appendix \ref{app:anal_perform_changes}, we analyse the changes in performance of the model fine tuned with the majority label compared to the base model.
We visualise the models' (FT (Maj.Label)) changes in performance against context open-endedness. We approximate that using the TVD between human oracles. We assume that a lower TVD, reflecting lower disagreement among human populations, indicates more `restrictive' contexts, while a higher TVD, indicates contexts that admit a higher level of plausible variability. 
We plot changes in performance by computing the differences between the TVD of the fine tuned model (FT Maj. label) and human CPD and the TVD of the base model and human CPD. Results are shown in \Cref{fig:differences_tvd_vs_tvdoracles_majlabel}. When comparing with the corresponding plots for FT (Mul.label) in \Cref{fig:differences_tvd_vs_tvdoracles}, we observe how improvements occur for contexts that admit lower plausible variabiltiy (\emph{i.e.} lower TVD among oracles values; steeper regression line towards lower Oracle TVD values for lower negative differences/performance gains). 

\section{Varying training labels per instance Study}\label{App:ablation}
\begin{figure}[t]
  \includegraphics[width=\columnwidth]{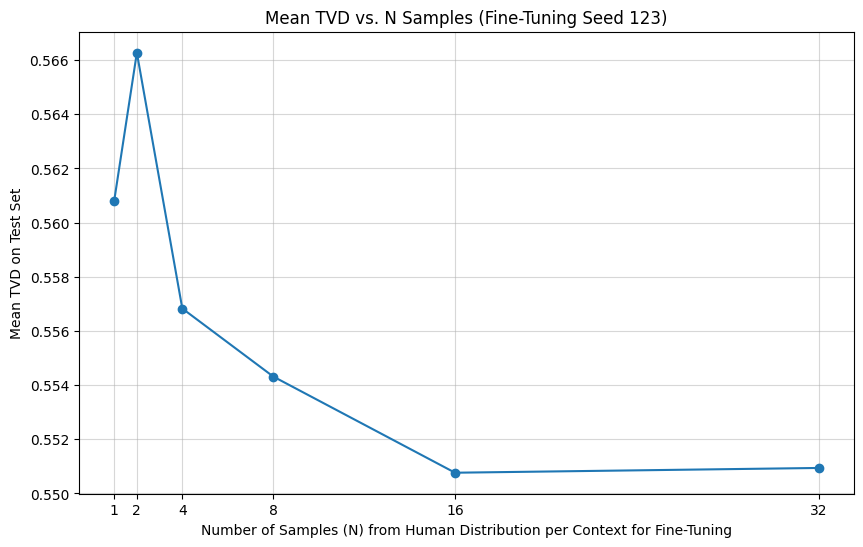}
  \caption{Mean TVD by number of samples per context. Performance improves with more samples, plateauing after 16.}
  \label{fig:ablation_samples}
\end{figure}

We perform an ablation to understand the number of labels that is necessary to obtain substantial performance gains. We perform this ablation study only for GPT2, given computational constrains (Mistral-7B-IT is a much larger model, and fine-tuning it repeatedly is computationally prohobited). We sample 1,2,4,16 and 32 labels given our available annotations and fine tune GPT2 given the subsequent training sets. We then perform the same evaluation as for the rest of our analysis and present the average TVD of the test set, against the label set size per instance in \Cref{fig:ablation_samples}.
Scores for 16 and 32 samples are nearly identical, and very similar to the score obtained when training on all available labels (40 on average per prompt). These results suggest that around 16 labels per instance are sufficient to observe significant performance gains. %while more human responses improve alignment with the empirical distribution, the marginal benefit diminishes, indicating that a limited number of high-quality annotations is sufficient to capture most human-like variability.

\section{Analysis for QA task without variability}
\label{App:dataset}
Whereas optimising for a task that admits inherent variability (\emph{i.e.} next-word prediction) might improve the model's ability to reproduce such variability better; the effect of this on tasks that admit no variability is unclear. We test the models' performance on knowledge-based question answering (which is a task that admits no plausible variability), adapted as a next-word prediction task. We create a small evaluation dataset based on a knoweldge-based question answering dataset, WebQuestions \citep{berant-etal-2013-semantic}.
We create a subset of 55 handpicked contexts, chosen to include a variety of topics ranging from science, history and pop culture, each rephrased into next-word prediction tasks. We demonstrate 3 randomly chosen examples below:
% \textbf{Prompt:}
% \begin{verbatim}
%     Instruction: Return one plausible next 
%     word for the following context. 
%     Context: The  first country to invade 
%     poland in ww2 was
%     Continuation: 
    
%     Instruction: Return one plausible next 
%     word for the following context. 
%     Context: the organelle responsible 
%     for atp production and storage is the
%     Continuation: 

%     Instruction: Return one plausible next 
%     word for the following context. 
%     Context: darth vader's star 
%     destroyer was called
%     Continuation: 
% \end{verbatim}
\textbf{Prompt:}
\begin{verbatim}
    Instruction: Return one plausible next 
    word for the following context. 
    Context: The  first country to invade 
    poland in ww2 was
    Continuation: 
\end{verbatim}

\textbf{Target:}
\begin{verbatim}
    Germany
\end{verbatim}

\textbf{Prompt:}
\begin{verbatim}
    Instruction: Return one plausible next 
    word for the following context. 
    Context: the organelle responsible 
    for atp production and storage is the
    Continuation: 
\end{verbatim}

\textbf{Target:}
\begin{verbatim}
    mitochondrion
\end{verbatim}

\textbf{Prompt:}
\begin{verbatim}
    Instruction: Return one plausible next 
    word for the following context. 
    Context: darth vader's star 
    destroyer was called
    Continuation: 
\end{verbatim}

\textbf{Target:}
\begin{verbatim}
    Devastor
\end{verbatim}

We also evaluate model performance using the original questions.  For Mistral-7B-IT, the instruction was modified into:
\textbf{QA-Prompt:}
\begin{verbatim}
    Instruction: Answer the following
    question with one word only 
    Context: What country first invaded 
    poland in ww2?
    Continuation: 
\end{verbatim}

We compare the base model, the model fine tuned with the original corpus (so as to account for the impact of training on Provo corpus, a potentially different domain) and the model that was fine tuned with multiple labels in their ability to generate the correct answer to the question (phrased as a next-word prediction task). 
To evaluate the performance, for each context, we sample 40 responses and measure how often responses exactly match the reference, denoted as hit rate.

\begin{figure*}[t]
  \centering
  \begin{subfigure}[t]{0.48\textwidth}
    \includegraphics[width=\textwidth]{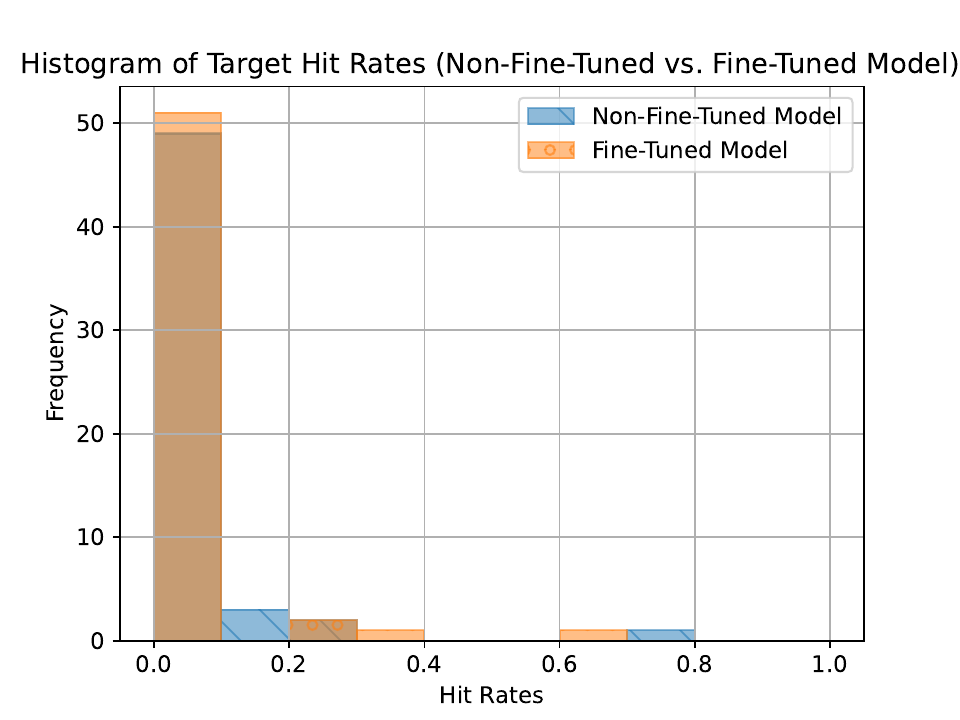}
    \caption{GPT-2}
  \label{fig:hit-rates-comparison_gpt_qa_corpus}  \end{subfigure}
  \hfill
  \begin{subfigure}[t]{0.48\textwidth}
    \includegraphics[width=\textwidth]{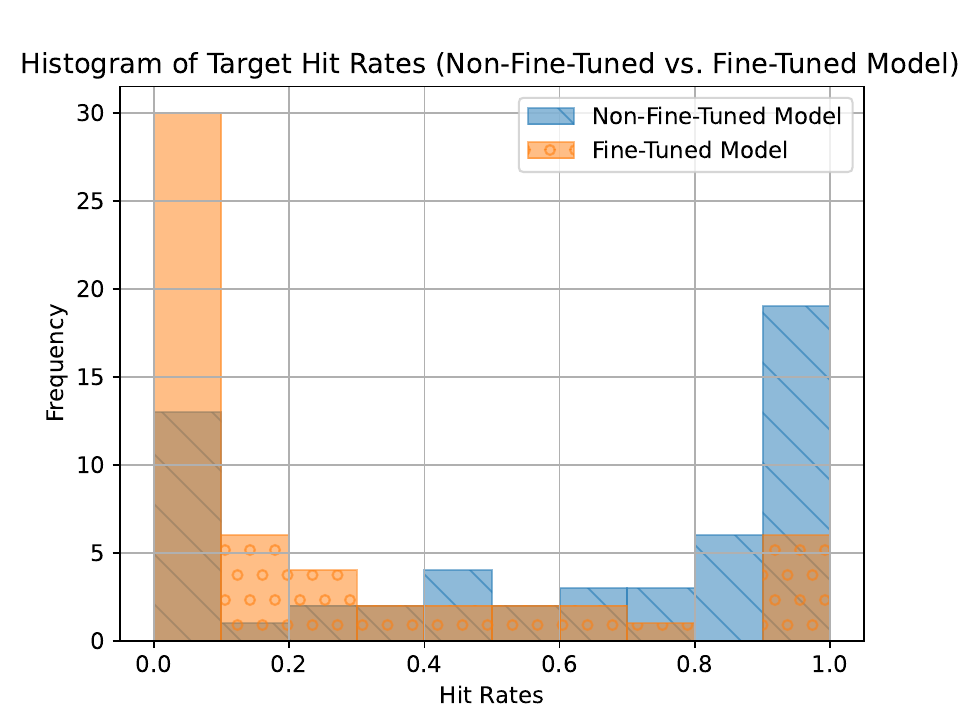}
    \caption{Mistral-7B}
    % \label{fig:hit-rates-comparison_it_qa}
  \end{subfigure}
\caption{Hit rates on gold target label after finetuning on hard targets (corpus). Averaged across 3 seeds.}
  \label{fig:hit-rates-comparison_corpus_ht_corpus}
\end{figure*}

Table~\ref{tab:hit_rate_results} shows the results of this evaluation.  %Fine-tuned models often do not return the target at all, in stark contrast to their non-finetuned counterparts. 
GPT-2 shows a slight increase in hit rate after finetuning, although its overall performance remains poor, and Mistral-7B-IT's performance also drops, more substantially.
However, we cannot rule out the effect of other confounders in the data or optimisation process that might have incidentally impacted the performance changes and are not relevant to the multiplicity of responses. Hence, we approach these preliminary results with cautiousness, and hope to inspire future work that investigates this more extensively. Supplementary histograms of hit-rates across contexts can be seen in \Cref{fig:hit-rates-comparison_corpus_ht_corpus}.
%We also show the results of finetuning on the original corpus continuations as a baseline for finetuning on Provo Corpus. 
% This may be attributed to GPT-2's limited model capacity and comparatively smaller knowledge base, which hinders its ability to retain factual information. 
% The performance drop observed for Mistral-7B aligns with the increased output variability illustrated in Figure~\ref{fig:entropy}, highlighting the trade-off between diversity and precision in tasks that rely on a single correct answer.

\begin{table}
  \centering
  \scalebox{0.9}{
  \begin{tabular}{l cc}
    \hline
    \multicolumn{3}{c}{\textbf{Mean Hit Rate $\pm$ SD}} \\
    \textbf{Model} & \textbf{GPT-2} & \textbf{Mistral-7B} \\
    \hline
    Base   & 0.032 $\pm$ 0.002 & 0.590$\pm$0.005 \\
    FT (Orig.corpus)  & 0.030 $\pm$ 0.002 & 0.229$\pm$0.005 \\
    FT (Mul.labels)       & 0.041 $\pm$ 0.002 & 0.127$\pm$0.005 \\
    \hline
  \end{tabular}}
  \caption{Mean target hit rate for 40 samples per context across three seeds with standard deviation, for both GPT-2 and Mistral-7B.}
  \label{tab:hit_rate_results}
\end{table}

\begin{figure*}[t]
  \centering
  \begin{subfigure}[t]{0.48\textwidth}
    \includegraphics[width=\linewidth]{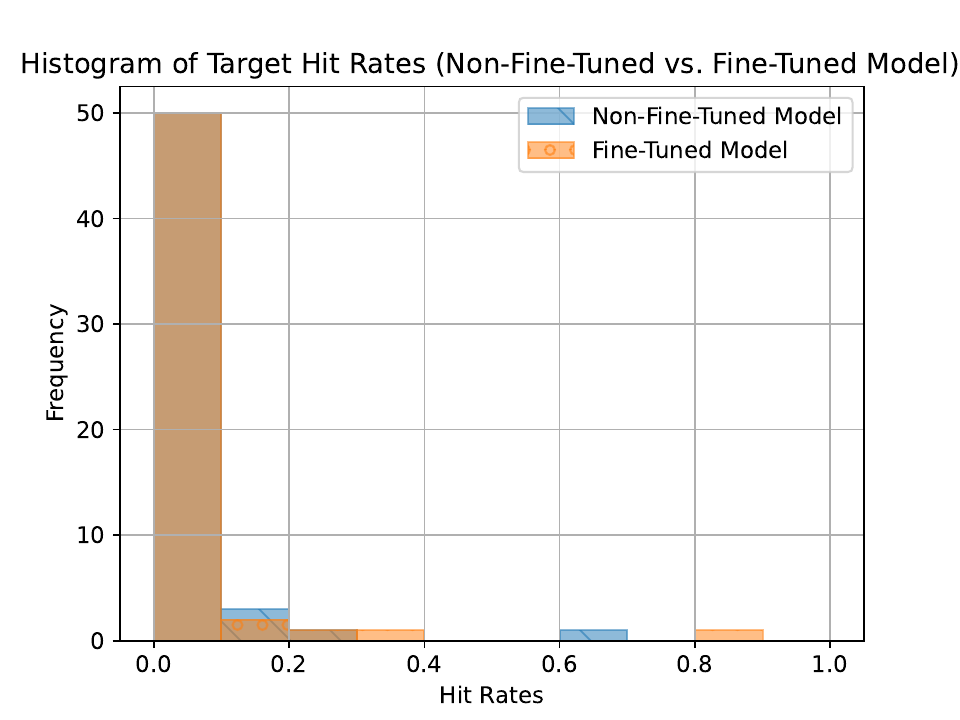}
    \caption{GPT-2}
  \label{fig:hit-rates-comparison_gpt}  \end{subfigure}
  \hfill
  \begin{subfigure}[t]{0.48\textwidth}
    \includegraphics[width=\linewidth]{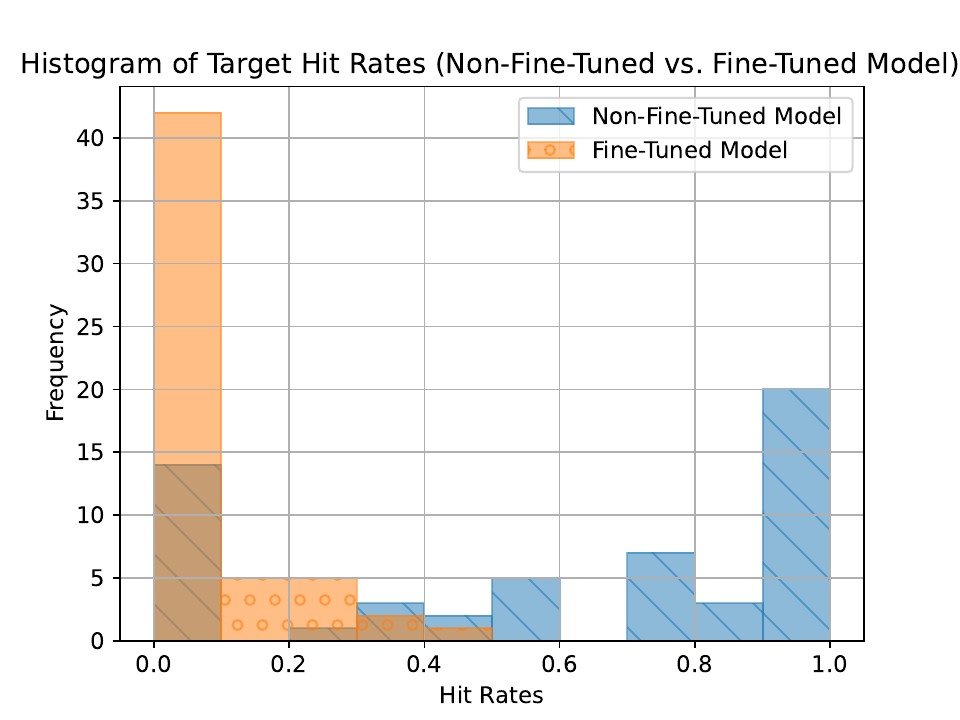}
    \caption{Mistral-7B}
    \label{fig:hit-rates-comparison_it}
  \end{subfigure}
  \caption{Hit rates on gold target label before and after finetuning. Averaged across 3 seeds.}
  \label{fig:hit-rates-comparison}
\end{figure*}

\begin{figure*}[t]
  \centering
  \begin{subfigure}[t]{0.48\textwidth}
    \includegraphics[width=\textwidth]{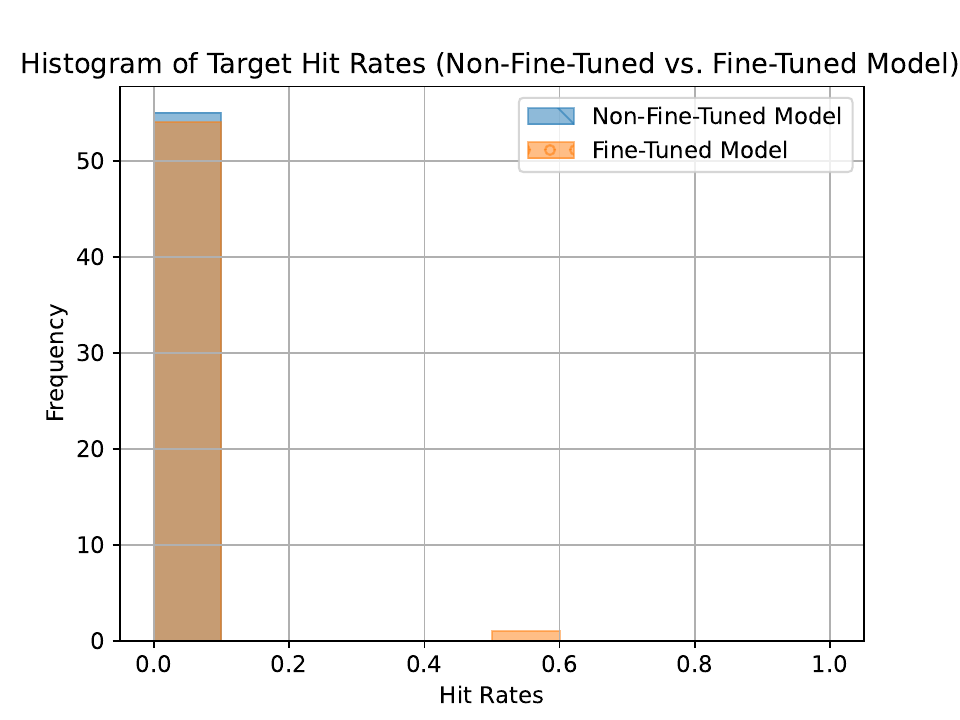}
    \caption{GPT-2}
  \label{fig:hit-rates-comparison_gpt_qa}  \end{subfigure}
  \hfill
  \begin{subfigure}[t]{0.48\textwidth}
    \includegraphics[width=\textwidth]{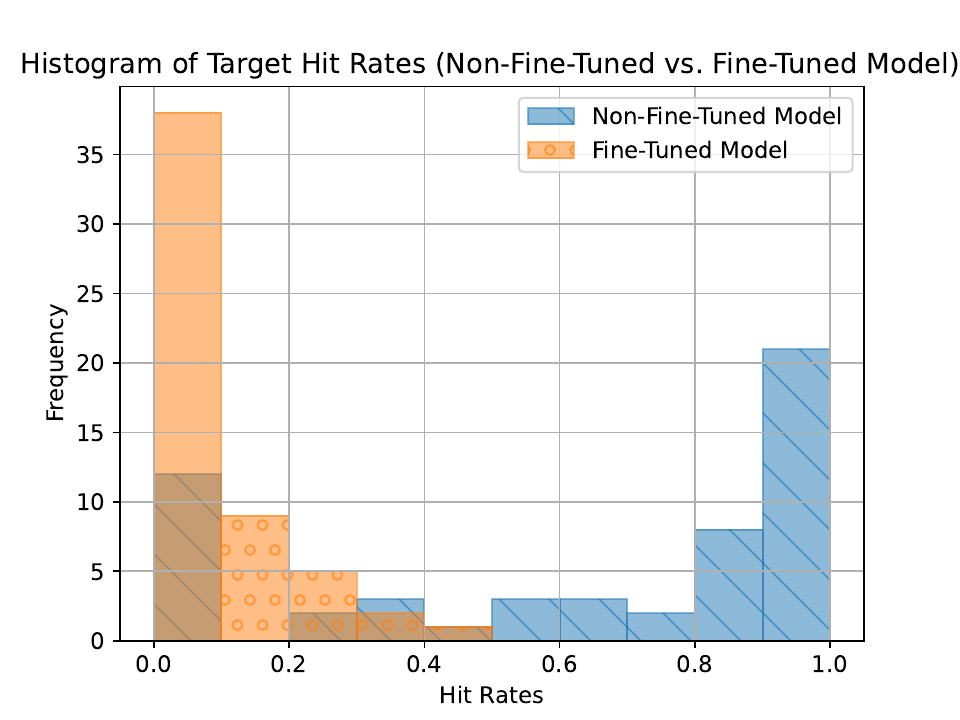}
    \caption{Mistral-7B}
    \label{fig:hit-rates-comparison_it_qa}
  \end{subfigure}
\caption{Hit rates on gold target label when prompted in the original QA format, before and after finetuning. Averaged across 3 seeds.}
  \label{fig:hit-rates-comparison_QA}
\end{figure*}

% \begin{figure*}[!h]
% \centering
% \includegraphics[width=\textwidth]{imgs/entropy_comparison.png}
% \caption{Shannon entropy of output distribution on single ground-truth data presented as a next word prediction task.}
% \label{fig:entropy}
% \end{figure*}

% \begin{figure*}[!h]
% \centering
% \includegraphics[width=\textwidth]{imgs/entropy_comparison_QA.png}
% \caption{Shannon entropy of output distribution on single ground-truth data presented as question-answerering task.}
% \label{fig:entropy_QA}
% \end{figure*}
\end{document}